\documentclass{article} % For LaTeX2e

\PassOptionsToPackage{numbers, compress}{natbib}
\usepackage[preprint,main]{neurips_2026}

\usepackage{amsmath,amsfonts,bm}

\def\eqref#1{equation~\ref{#1}}
\def\1{\bm{1}}

\DeclareMathAlphabet{\mathsfit}{\encodingdefault}{\sfdefault}{m}{sl}
\SetMathAlphabet{\mathsfit}{bold}{\encodingdefault}{\sfdefault}{bx}{n}

\DeclareMathOperator*{\argmax}{arg\,max}

\usepackage{hyperref}
\usepackage{tcolorbox}
\usepackage{url}
\usepackage{amsmath}
\usepackage{graphicx}
\usepackage{amssymb}
\usepackage{makecell}
\usepackage{multirow}
\usepackage{wrapfig}
\usepackage{booktabs}

\title{Integrating Simulation and Chain-of-thought Reasoning in Multimodal-Language Models For Physical Reasoning}

\author{
Yingqiao Wang\thanks{Corresponding author.}\\
Department of Psychology, Harvard University\\
Sakana AI\\
\texttt{yingqiaowang@g.harvard.edu}
\And
Eric Bigelow\\
Department of Psychology, Harvard University\\
Goodfire AI\\
\texttt{ebigelow@g.harvard.edu}
\And
Tomer Ullman\\
Department of Psychology, Harvard University\\
\texttt{tullman@fas.harvard.edu}
\AND
Yujin Tang\\
Sakana AI\\
\texttt{yujintang@sakana.ai}
\And
Sebastian Risi\\
Sakana AI\\
\texttt{sebastianrisi@sakana.ai}
}

\begin{document}

\maketitle

\begin{abstract}

Multimodal language models often need to reason about the physical dynamics of the world in order to answer questions appropriately.
Various methods have emerged for solving physical reasoning tasks, and in particular, two leading approaches are Chain-of-Thought reasoning and physics-grounded simulation.
However, relatively little is known about which particular problems can be solved by each approach. Some tasks may benefit from computationally expensive generative video simulations, whereas others may be solved equally well (or even better) with language-based reasoning.
Our work addresses this by systematically comparing simulation and language reasoning across a variety of multimodal tasks involving physical reasoning, such as answering questions about fluid dynamics and elasticity. 
We draw a parallel with this to resource-rational theories of human cognition, which propose that people optimally trade off simulation and reasoning depending on task demands.
We present a novel cognitively-inspired hybrid model which optimally trades off simulation and reasoning to improve model performance and computational efficiency.
Our model is able to decide the appropriate method to solve a physical reasoning problem at test time, and finds an optimal balance between accuracy and cost. Further, our model can be trained on a smaller vision-language model that is only 4B parameters and deployed to improve performance on much bigger base models with $>$3T parameters like Gemini 3 Pro. 

\end{abstract}

\section{Introduction}

Physical reasoning is an important topic in the current AI community, and it is needed to overcome important challenges like ensuring that embodied AI agents act safely in the world \citep{zheng2024contphy} \citep{chow2025physbench}. There are different methods to implement such reasoning ability. Chain-of-Thought \cite{kojima2022large} \cite{wei2022chain} is one dominant method, used to solve many reasoning tasks, including physical reasoning. The recent SoTA physical reasoning models like \cite{cosmos_cookbook_av_captioning_vqa_2026} adapted the chain-of-thought as the only reasoning methods when reasoning about physical tasks. However, this might not be the only method to implement physical reasoning ability. 

In cognitive science, an analogous line of research is called intuitive physics, which studies how human approaches physical reasoning. Current approaches suggests that mental simulation could be a way that people approach physical reasoning \citep{battaglia2013simulation} \citep{ullman2017mind}. Researchers in AI had also adapted this idea of mental simulation into grounding the physical reasoning in Multi-modal Large Language Models (MLLMs) \citep{liu2022mind}. Furthermore, recent breakthrough in diffusion model-based video model has shown emerging zero-shot ability as a physics simulator \citep{wiedemer2025video} and it's ability to push towards intelligence in robotics \citep{veorobotics2025}. 

In theory, Chain-of-thought is considered to be computationally cheaper when compared with simulation with physics simulator, but simulation is considered to be more accurate since it is grounded on real physics . This raises an interesting observation that whether each method would be better suited to solve specific types of physical reasoning tasks, thus creating this unique tradeoff between accuracy and computational costs in physical reasoning in MLLMs when different method was used.

Such dynamical selection based on tradeoff between accuracy and computational costs is actively adapted in areas like mathematical reasoning \citep{waheed2025less} and LLM Agentic Reasoning \citep{yang2026ares}. Given that reasoning effort is tunable in LLMs equipped with Chain-of-thought, these work have found that dynamically adjusting the reasoning effort based on the questions the model is facing could not only increase accuracy, but also manage computational cost to prevent wasting computational power. Furthermore, works like \citep{liu2024mind} have found that reasoning more does not help with all the problems. Over-reasoning might hurt the performance of the LLM on certain types of problems, and this finding aligns with human studies. 

Therefore, to answer the questions that whether such dynamical method selection between simulation and chain-of-thought based on tradeoff between accuracy and computational costs applies to physical reasoning, we propose a model that not only empirically shows that such tradeoff between accuracy and computational costs exists, but also able to find the best tradeoff between accuracy and computational costs on various physical reasoning tasks, inspired by research in resource rationality in cognitive science. 

% dynamic reasoning effort adjustment in LLM

\section{Resource Rationality in Intuitive Physics}

\label{rational}

People are able to efficiently reason about the physical dynamics of everyday objects. For example, if you saw a ball flying towards you, you might quickly predict where to catch the ball. This ability of people being able to reason about physics of the world is called intuitive physics. Many competing theories have been proposed to explain this `intuitive physics', and recent research have suggested that it is likely that humans use a combination of different methods to carry out this reasoning \cite{hartshorne2025insights}. Interestingly, previous work in cognitive science have suggested a hybrid model that combines two most competing methods, which are mental simulation and heuristic reasoning \citep{smith2023integrating} \citep{sosa2025blending}. 

The hybrid models are designed around one core concept: resource rationality. In \cite{smith2023integrating}, they proposed the Integration of Simulation and Rules (ISR) framework based on resource rationality before performing a task. Before performing physical reasoning on a specific task $T$, we have to decide on a strategy $S$ to answer this task. But, several factors that are associate with this strategy needs to be balanced and considered. There are mainly two factors to consider in the ISR: the metabolic costs $C$ for adapting this strategy, and the expected utility $U$ of this strategy, needs to be considered. Therefore, the ISR framework aims to optimize for the best strategy $S^*$ that uses the resource in the most rational way as the following:
\begin{equation}
    S^* = \argmax E(U(S,T) - C(S,T))
\end{equation}
Furthermore, in \cite{sosa2025blending}, a hybrid model that could be applied when performing a task was proposed. When humans are performing physical reasoning, even though it is possible that they are
adapting a simulation strategy when doing a physical reasoning task, some parts in the task that can be resolved by using heuristics rules could be used to save cognitive resource, achieving resource rationality.
This idea of resource rationality in intuitive physics of human is useful for many cases, and serves as a motivation for our model. Resource rationality has been shown to be useful in intuitive physics for reducing the amount of computation humans need to perform for physical reasoning, while remaining accurate. This serves as inspiration for our model, which similarly trades off physical simulation against heuristic-based reasoning, to improve model accuracy while also reducing costs."

\section{Hybrid Simulation and Reasoning Model}

\label{sec:model}

\begin{figure}[h!]
    \centering
    \includegraphics[width=\columnwidth]{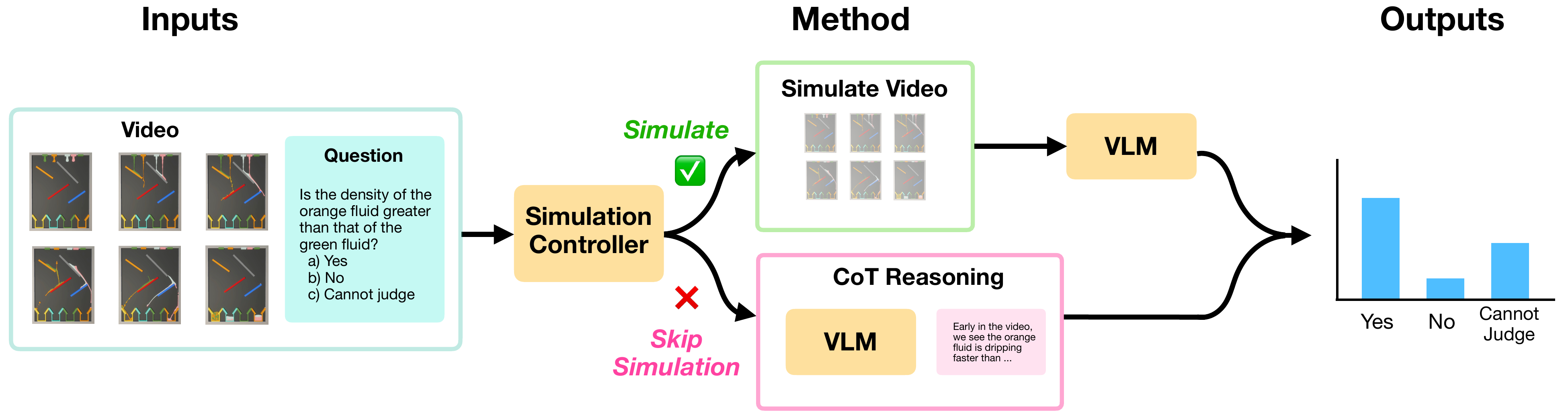}
    \caption{The overview of HSR model. Our HSR model consists of simulation controller, physics simulator-based simulation, and Chain-of-thought reasoner. The inputs consist of a set of video and a question based on this video, which is passed into simulation controller. Depending on the assignment from the simulation controller, either simulation or Chain-of-thought reasoner will be applied to answer the question. 
}
    \label{fig:introfig}
\end{figure}

% insert connections to Kevin's model
Here we present our model Hybrid Simulation and Reasoning Model (HSR) to incorporate resource rationality into physical reasoning.  Consider an arbitrary physical reasoning task that consist of a problem $s$ and a video $d$ that the problem is based on, and two methods of solving physical reasoning tasks: simulator and Chain-of-thought reasoning. First, we determine which of the two methods is best suited to solve this physical reasoning tasks by using a trained simulation controller. Second, we pass the physical reasoning tasks to the method determined by the simulation controller to get an answer for this physical reasoning tasks.

\subsection{Model Architecture}
We first process the problem $s$ and the video $d$ into embeddings. For visual embedding, we only use the first frame $x_1$ of the video $d$. We used visual embedding model and text embedding model to generate the corresponding embeddings. Therefore, we will be able to obtain a concatenated embedding $\mathbf{h}$ by:
\begin{equation}
\begin{aligned}
    \mathbf{v}=f_{visual}(x_1) \in \mathbb{R}^{512} \qquad \mathbf{u}=f_{text}(s) \in \mathbb{R}^{718} \qquad
    \mathbf{h} = [\mathbf{u}; \mathbf{v}] \in \mathbb{R}^{1230}
\end{aligned}
\end{equation}
We pass the concatenated embedding $\mathbf{h}$ into the simulation controller, as shown in fig \ref{fig:introfig}. The simulation controller is a trained MLP that that process the embedding $\mathbf{h}$ into a logit. We use a 2-hidden-layer MLP ($1230\rightarrow512\rightarrow256\rightarrow1$) with BatchNorm ($512,256$), ReLU, and dropout ($p=0.5$) after each hidden layer to produce a scalar logit $\ell$.

Where $\ell$ is the predicted logit from the model. The final predicted probability $p$ can be recovered by applying sigmoid function on logit: $$p = \sigma(\ell) = \frac{1}{1+e^{-\ell}}$$ 

During the training phase, the model aims to minimize the BCE With Logit Loss, stated as the following:
\begin{equation}
\begin{aligned}
\mathcal{L}_{\mathrm{BCE\text{-}logit}}(\ell,y)
= -\Big( y \log p + (1-y)\log(1-p) \Big)
\end{aligned}
\end{equation}
\subsection{Definition of Ground Truth Probability}
During the training phase, ground truth probability $y$ is provided for all data in the training set and is generated by the base VLM. Given that $y \in [0,1]$, let $(\Omega,\mathcal{F},\mathbb{P})$ be a probability space. Based on this probability space, we define a random variable $X:\Omega \to \{S,C\}$ where $S$ denotes using simulation and $C$ denotes using chain-of-thought only. 

Let $0 \le p_{\mathrm{sim}},p_{\mathrm{cot}} \le 1$. Define $p_{\mathrm{sim}}$ and $p_{\mathrm{cot}}$ to be the probability of using simulation and the probability of using chain-of-thought only reasoning. For each data entry in the training data, it was answered by using simulation K times, and Chain-of-thought reasoning K times. Therefore:
\begin{equation}
    \begin{aligned}
        p_{\mathrm{sim}} = \frac{1}{K}\sum^K_{i=1} \mathbb{I}\!\left[a_{\mathrm{sim}}^{(i)}=a_{\mathrm{gt}}\right] \in [0,1] \qquad
        p_{\mathrm{cot}} = \frac{1}{K}\sum^K_{i=1} \mathbb{I}\!\left[a_{\mathrm{cot}}^{(i)}=a_{\mathrm{gt}}\right] \in [0,1]      
    \end{aligned}
\end{equation}
where $a_{\mathrm{sim}}^{(i)}$ is the answer output by simulation method in $i$th attempt, $a_{\mathrm{cot}}^{(i)}$ is the answer output by Chain-of-thought method in the $i$th attempt, and $a_{\mathrm{gt}}$ is the ground truth answer. The indicator function allows partially correct answer.

We hereby define the ground truth probability $y$ as:
\[
y \triangleq \mathbb{P}(X=S)
= \frac{p_{\mathrm{sim}}}{p_{\mathrm{sim}}+p_{\mathrm{cot}}},
\]
\subsection{Method Selection Threshold}
We define a threshold $\tau_t$ to be $\tau_t = \tau_{accuracy} + \tau_{cost}$. All parameters $\tau_t$, $\tau_{accuracy}$, $\tau_{cost}$ are tunable hyperparameter, but $\tau_{accuracy}$ is determined by how the ground truth probability $y$ is defined and constructed, meaning that it is usually set as 0.5. $\tau_{cost}$ serves as a penalty term that penalizes the use of simulation due to the high metabolic cost of simulation. 

If $p\le\tau_t$ we assign the data entry as Chain-of-thought reasoning, and if $p > \tau_t$ then we assign the data entry as simulation. Notice that this threshold controls the number of data entries being assigned as simulation, thus controls the computational costs, since simulation is by definition more costly. 

Then, as shown in fig \ref{fig:introfig}, we pass the question and video into simulation or Chain-of-thought reasoning, depending on the decision made by the simulation controller to obtain the final answer to the question. In simulation, a fine-tuned video diffusion model is used as the physics-grounded simulator. We utilize a VLM for Chain-of-thought reasoning and readout phase in simulation.

\subsection{Design of Video Simulator}

\label{video-simulator}

In our HSR model, we implemented a compositional structure for the physics simulator we used in our model in order to make it support counterfactual simulation. First, a base VLM will takes a video and question and determines whether counterfactual simulation is needed. Then, if needed, the base VLM will propose the plan to construct the physics prior (conditional frames) needed for the video simulator to generate the video. In this plan, 1: The question will be rewritten to their non-counterfactual version and 2: the modifications will be suggested and passed to an image generation model for the modification. Then, the image generation model will modify the physics prior, and this physics prior will be passed to the physics simulator as the conditional prior for video generation. For the prompt that is used during this phase, see Appendix \ref{prompt}.

\section{Experiments}

\subsection{Hypothesis}
We hypothesize that Chain-of-thought and Simulation should be applied to specific categories of physical reasoning tasks. And our HSR model is able to find the best scheme to assign questions from different categories to the best method suitable for answering this question, by considering the tradeoff between accuracy and computational costs. Furthermore, we hypothesize the simulation controller in our HSR model is irreplaceable by a Zero-Shot VLM-as-a-Judge, and such a simulation controller could be trained using training data generated by a smaller model and still generalizes to a larger model. 

\subsection{Dataset and Model Details}
We used tasks from PhysBench dataset \citep{chow2025physbench} and ContPhy dataset \citep{zheng2024contphy} as it contains numerous realistic physical reasoning tasks from variety of categories. For more details regarding the dataset, please refer to Appendix \ref{dataset}.

The visual embedding models we used is CLIP ViT-B/32 \citep{Radfordetal2021} and the text embedding model we used is all-MiniLM-L6-V2 \citep{reimers-2019-sentence-bert}. For the simulator used in simulation, we used LTX-Video \citep{hacohen2024ltx} as the physics-grounded simulator. The Nano-Banana-Pro and Gemini-3-Pro \citep{team2023gemini} is used as the image generation model and base VLM to generate physical prior needed for counterfactual simulation, as shown in \ref{video-simulator}. For the VLM that was used in Chain-of-thought reasoning method and as a readout VLM in the simulation method, we used Gemini 3 Pro, Qwen3-VL-4B-Thinking, and Qwen3-VL-2B-Thinking \citep{bai2025qwen3} as the VLM. The Gemini-3-Pro is accessed through official Google API interface, and the LTX-Video model, Qwen3-VL models were run on a single A100 for maximum of 3 days.

\subsection{Sampling Details}

We adapt $K=5$ as the sampling parameter for our ground truth probability $y$. By the design of our ground truth probability $y$, we set $\tau_{accuracy} = 0.5$. For the penalization term $\tau_{cost}$, we set $\tau_{cost} = 0.03$. When answering a question, the model will be rewarded a score of 1 if all the answers were selected. If only the part of the correct answers were selected, a score of 0.5 will be rewarded. If none of the correct answers were selected, a score of 0 will be rewarded. 

\begin{figure*}[t!]
    \includegraphics[width=\linewidth]{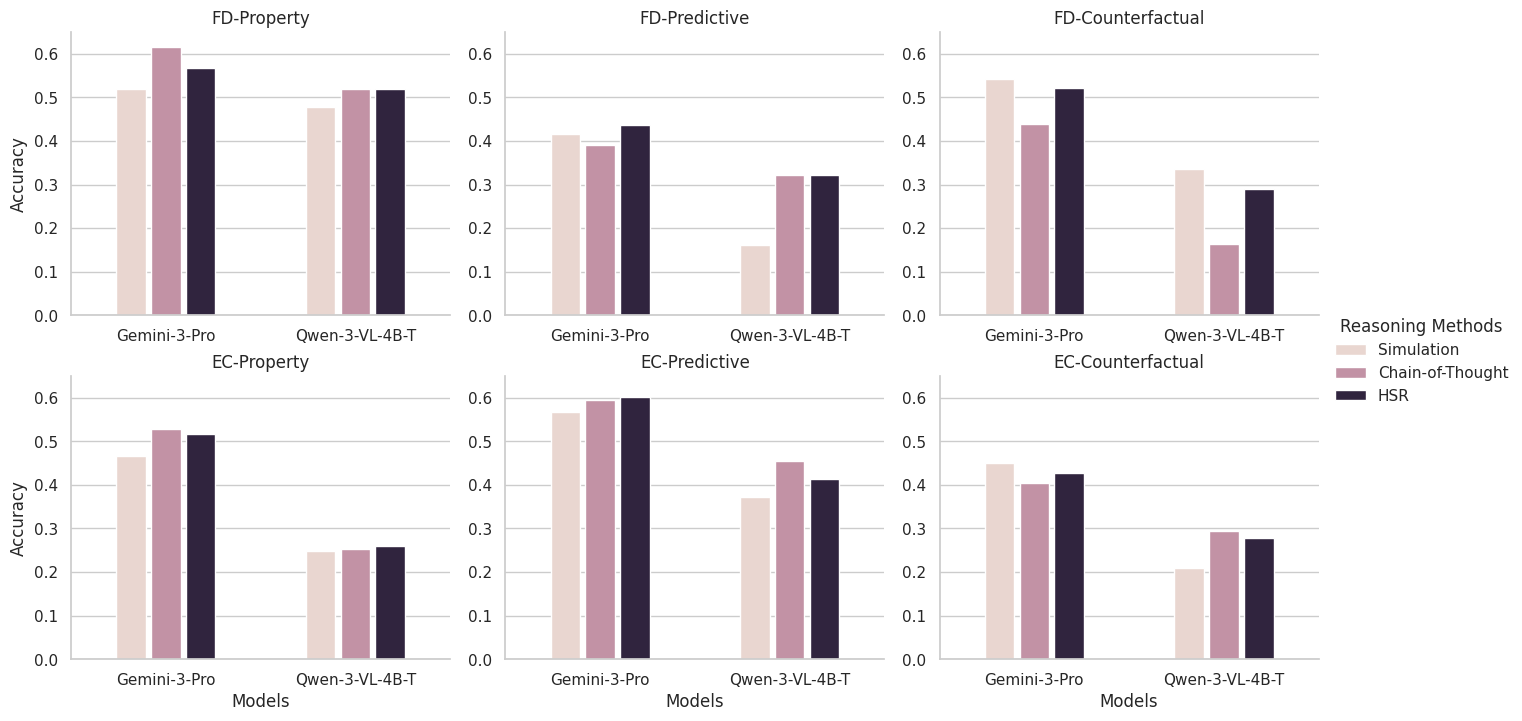}
    \caption{Evaluation results of simulation only (Simulation) and chain-of-thought only (Chain-of-Thought) when Gemini-3-Pro and Qwen3-VL-4B-Thinking was set as the base model. The evaluation is based on fluid hourglass (FD) and bouncing ball (EC) domains, with questions in property (Property), predictive (Predictive), and counterfactual (Counterfactual) categories. We could observe that HSR performs near the best accuracy across all physical domains and question categories}
    \label{table:evaluation}
\end{figure*}

\section{Results}

\subsection{Evaluating Chain-of-Thought and Simulation Across Various Physical Reasoning Tasks}

\paragraph{Accuracy in each question category differs by the method}Using Gemini-3-Pro and Qwen3-VL-4B-Thinking as the base VLM, we evaluated the performance of Chain-of-Thought and Simulation as two separate independent method across all question categories in two domains: fluid hourglass and bouncing ball and all categories: property, predictive, and counterfactual. As shown in table \ref{table:evaluation}, we observe that simulation is able to achieve greater accuracy in counterfactual category by a large margin. Noticably, in fluid hourglass - counterfactual (FD-Counter), the simulation outperformed Chain-of-Thought by 10\% when Gemini-3-Pro was used as the base model, and 17\% when Qwen3-VL-4B-Thinking was used as the base model. Conversely, simulation did not perform better than Chain-of-Thought in the property category. In categories like bouncing ball - property (EC-Prop), we observe a performance gap, where chain-of-thought outperformed simulation by 27\%. In addition, we see a relatively mixed signal in the predictive category, indicating that in this question category, simulation and chain-of-thought could serve as two methods that complement each other.
\label{category-difference}

\begin{wrapfigure}{r}{0.5\textwidth}
  \begin{center}
    \includegraphics[width=0.48\textwidth]{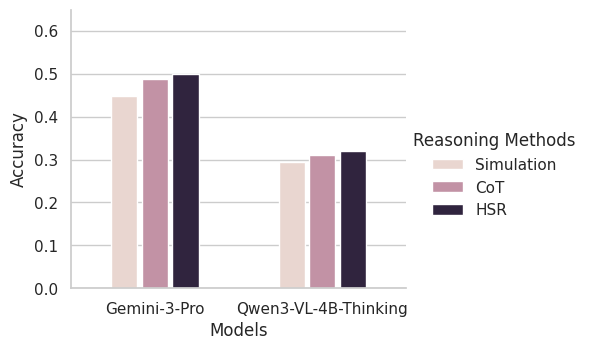}
  \end{center}
  \caption{Evaluation results of simulation only (Simulation) and chain-of-thought only (Chain-of-Thought) when Gemini-3-Pro and Qwen3-VL-4B-Thinking was set as the base model. The evaluation is based on overall performance  We could observe that HSR achieved the overall accuracy across all physical domains and question categories}
\label{fig:overall}
\end{wrapfigure}

% \begin{table}[h!]
% \centering
% \begin{tabular}{|p{1.5cm}|p{1.2cm}||p{1cm}|p{1cm}||p{1cm}|p{1cm}||p{1cm}|p{1cm}||p{1cm}|}
% \hline
% Base Model & Methods & FD-Prop & EC-Prop & FD-Pred & EC-Pred & FD-Counter & EC-Counter & Overall\\
% \hline
% \multirow{3}{1.5cm}{Gemini-3-Pro} & Sim & 0.519 & 0.252 & \textbf{0.363} & 0.568 & \textbf{0.541} & \textbf{0.451} & 0.449 \\
% \cline{2-9} 
% & CoT & \textbf{0.615} & \textbf{0.529} & 0.337 & 0.595 & 0.440 & 0.403 & 0.487
%  \\
% \cline{2-9}
% & HSR & 0.567 & 0.516 & 0.362 & \textbf{0.602} & 0.522 & 0.428 & \textbf{0.500} \\

% \hline
% \hline
% \multirow{3}{1.5cm}{Qwen3-VL-4B-Thinking}  & Sim & 0.478 & 0.248 & 0.119 & 0.373 & \textbf{0.335} & 0.208 & 0.2935 \\
% \cline{2-9}
% & CoT & 0.519 & \textbf{0.252} & \textbf{0.180} & \textbf{0.455} & 0.163 & \textbf{0.295} & 0.311 \\
% \cline{2-9}
% & HSR & \textbf{0.529} & 0.252 & 0.162 & 0.413 & 0.290 & 0.279 & \textbf{0.321} \\
% \hline
% \end{tabular}
% \label{table:evaluation}
% \end{table}

\subsection{Evaluating HSR Across Physical Reasoning Tasks}

\label{result:HSR}

As we demonstrated above, simulation and chain-of-thought does not achieve best accuracy across all physical domains and question categories, therefore, our HSR model is essential here as it is equipped with a simulation controller that routes questions in test-time to the best method that is suitable to solve this question. Here we will evaluate and analyze the model in depth.

\paragraph{HSR achieve best accuracies on average} As shown in the Figure \ref{table:evaluation} and \ref{fig:overall}, we also evaluated the performance of HSR under two physical domain and three question categories. We notice that HSR is able to achieve near-best performance across all physical domains and question categories, thus achieve the best 

average performance when Gemini-3-Pro and Qwen3-VL-4B-Thinking was set as the base VLM, comparing with just simulation or just chain-of-thought. This demonstrates the HSR's ability to perform at near-best accuracy regarding the question categories and physical domains, highlighting the importance of routing scheme and the HSR model.

Since our HSR model's accuracy is entirely based on base VLMs, which our HSR model provides only a routing scheme that assigns each question to a suitable method, including simulation and chain-of-thought, we performed additional analysis probing specifically the impact of base VLM ability on the HSR model. Please refer to Appendix \ref{additional} for further details.

\begin{figure}[t]
  \begin{minipage}[b]{0.33\textwidth}
    \includegraphics[trim=5 8 10 10,clip,width=\textwidth]{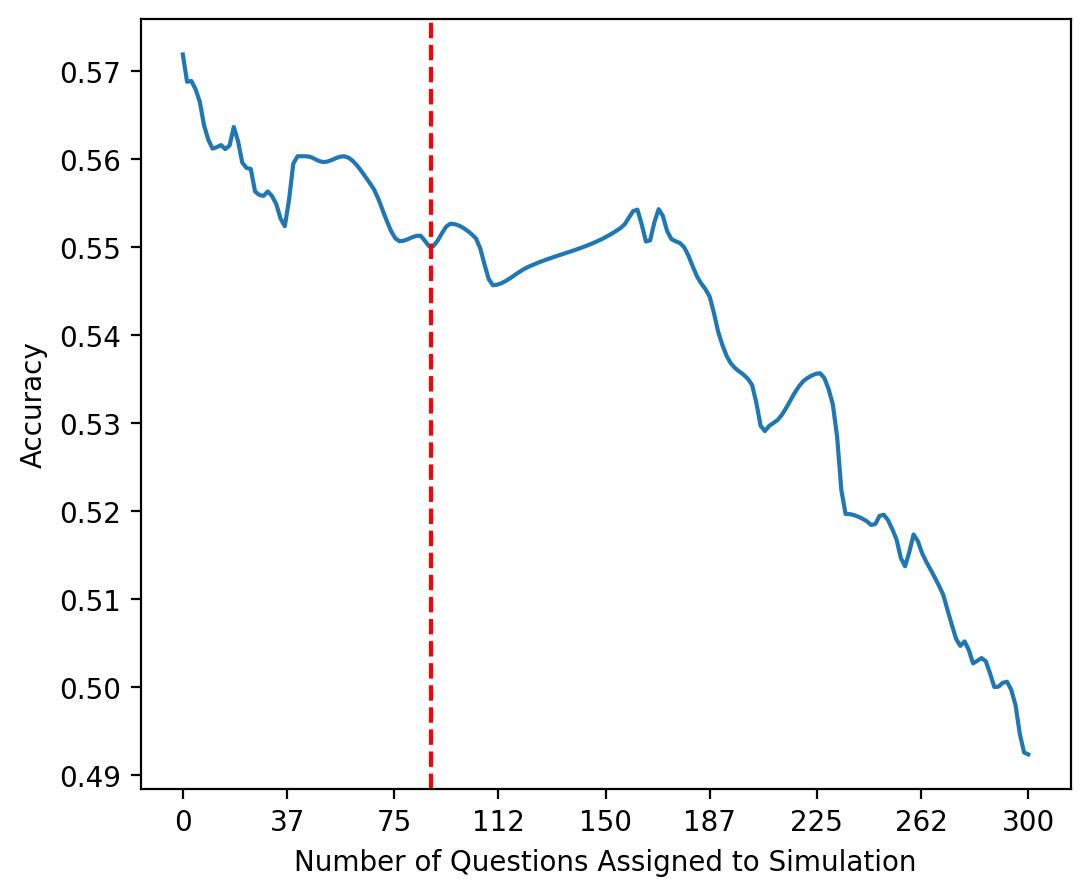}
  \end{minipage}
  \begin{minipage}[b]{0.33\textwidth}
    \includegraphics[trim=5 8 10 10,clip,width=\textwidth]{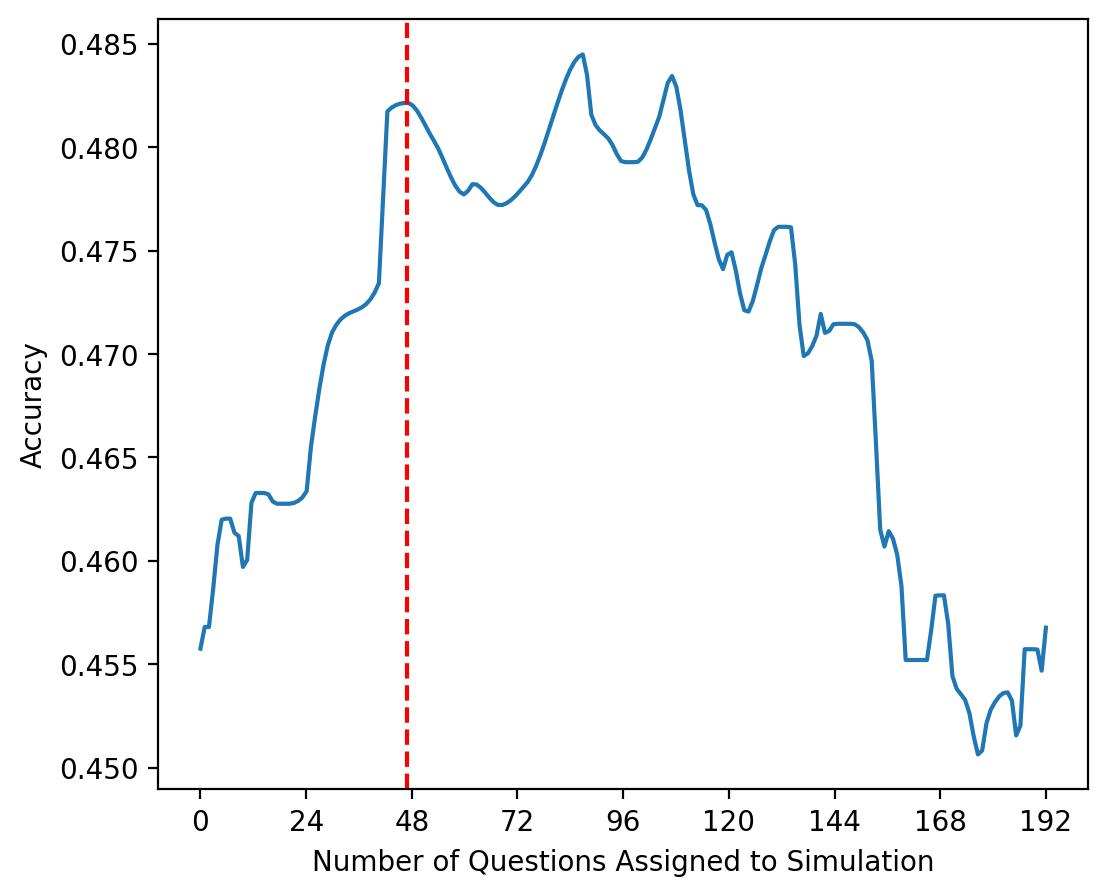}
  \end{minipage}
  \begin{minipage}[b]{0.33\textwidth}
    \includegraphics[trim=5 8 10 10,clip,width=\textwidth]{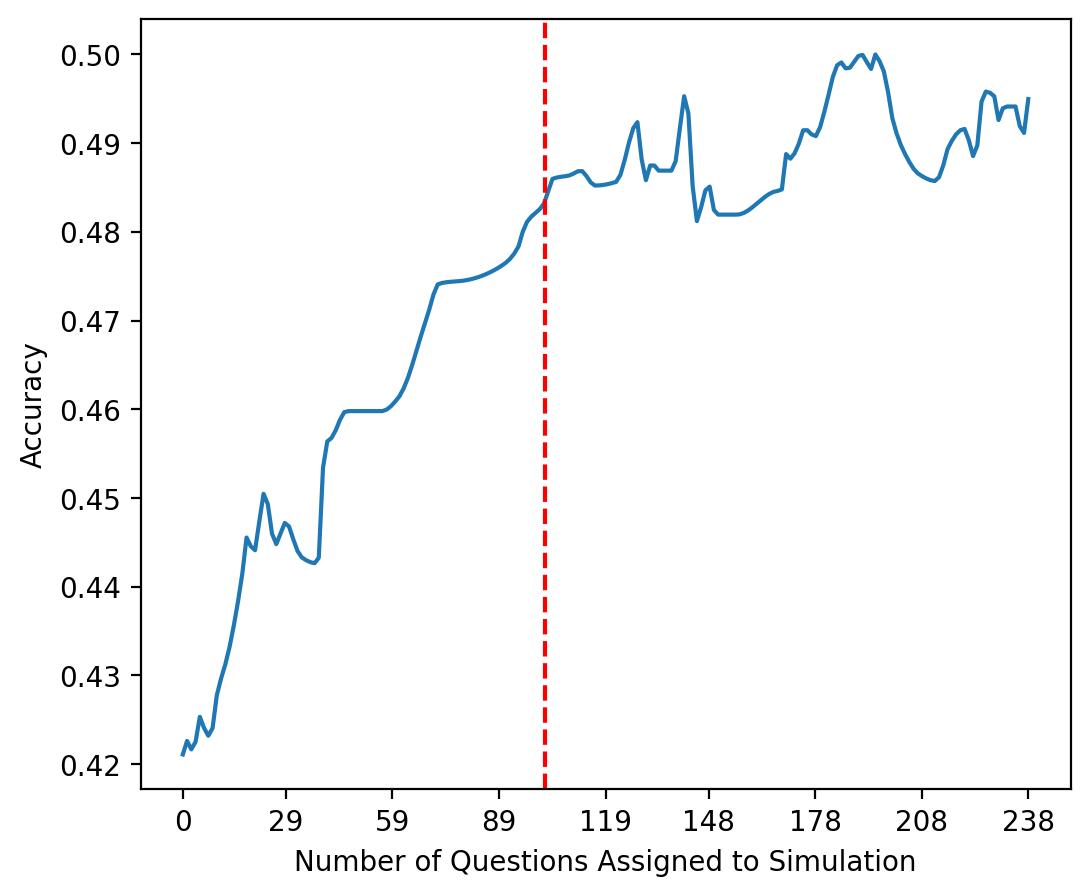}
  \end{minipage}
  \\[0.5em]
  % Second row
  \begin{minipage}[b]{0.33\textwidth}
    \includegraphics[trim=5 8 10 10,clip,width=\textwidth]{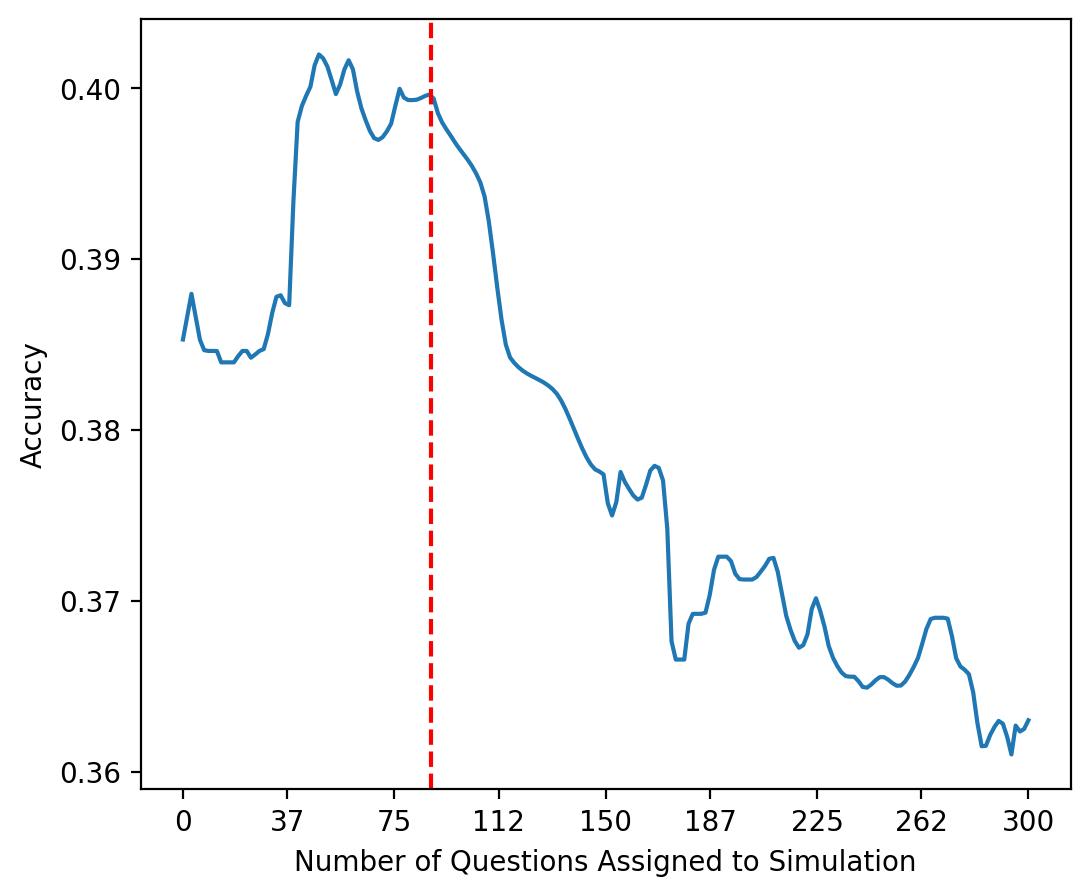}
  \end{minipage}
  \begin{minipage}[b]{0.33\textwidth}
    \includegraphics[trim=5 8 10 10,clip,width=\textwidth]{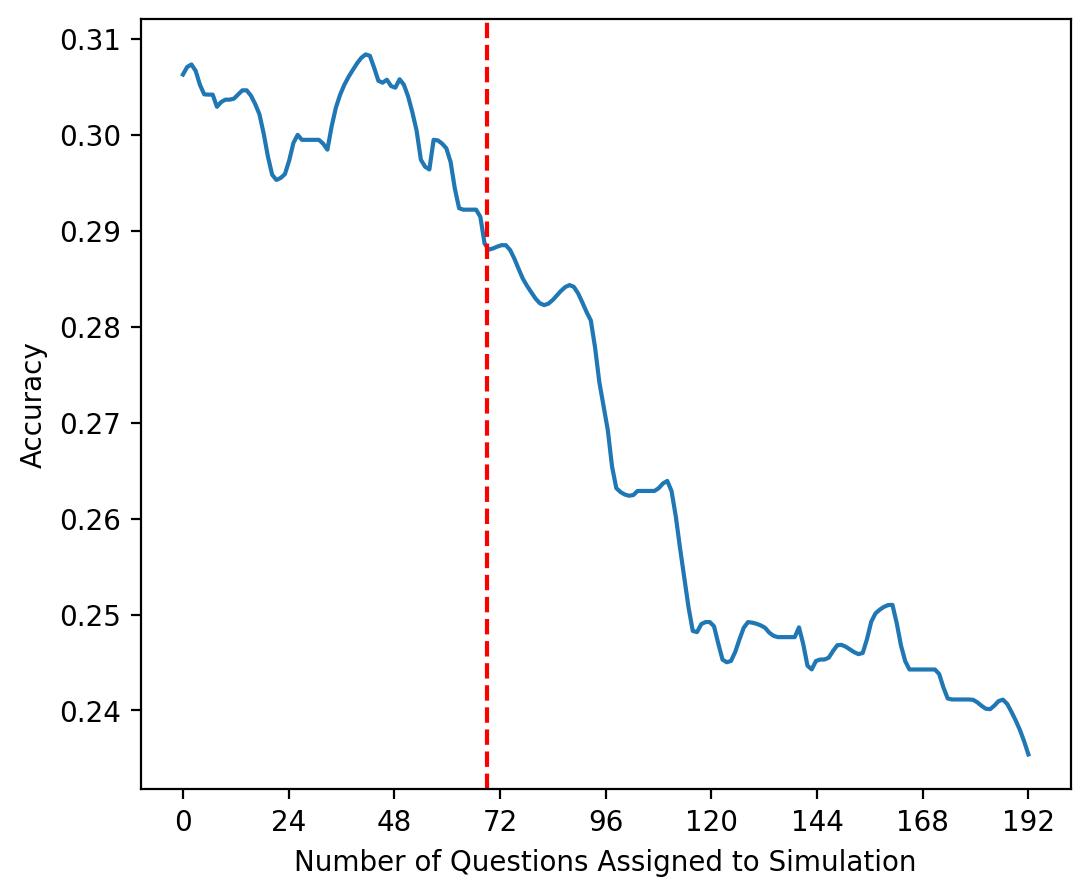}
  \end{minipage}
  \begin{minipage}[b]{0.33\textwidth}
    \includegraphics[trim=5 8 10 10,clip,width=\textwidth]{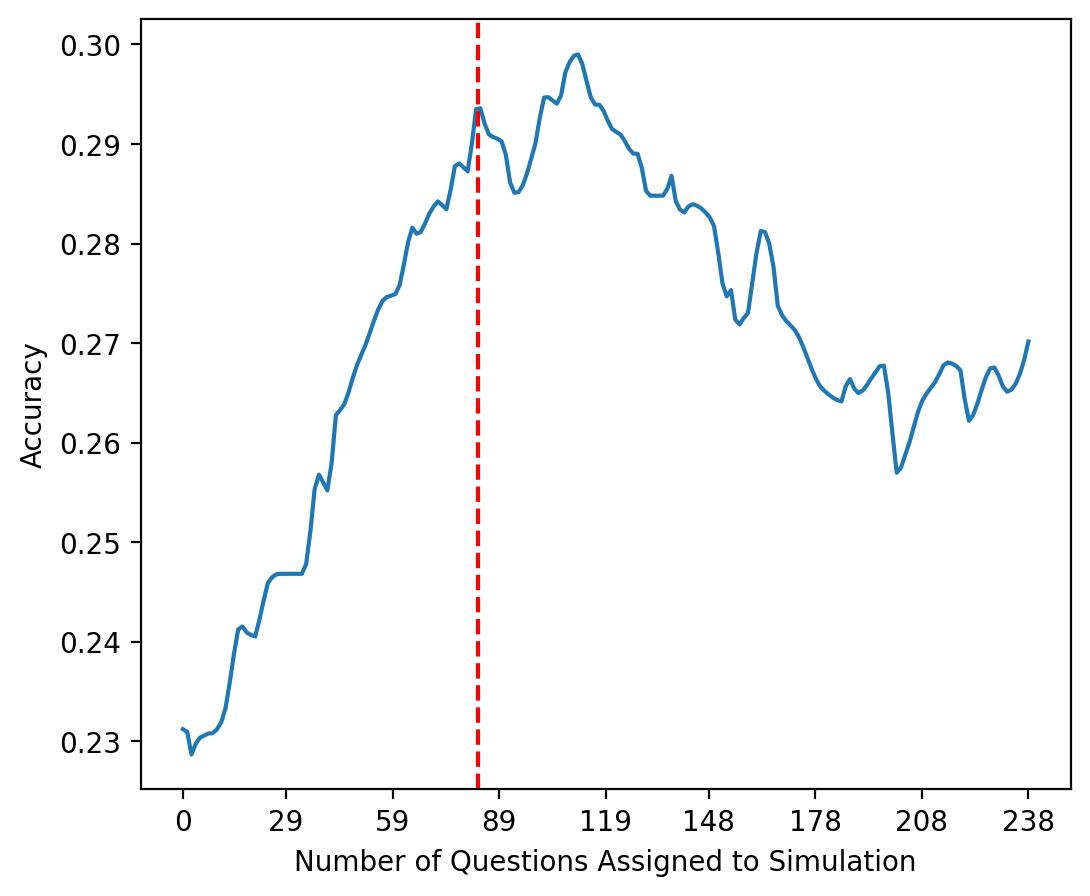}
  \end{minipage}
  \label{fig:model-results}
  \caption{Accuracy vs Number of Questions Assigned to Simulation. The first row is the HSR model with Gemini-3-Pro as the base model. The second row is the HSR model with Qwen3-VL-4B-Thinking as the base model. The first column is the property category collasped across two physical domains. The second column is the predictive category collapsed across two physcial domains. The third column is the counterfactual category collapsed across two physical domains. The red dash line is the number of questions assigned to simulation, which is determined by HSR model at threshold $\tau_t = 0.53$. This dash line also indicates the local maximum of the function, indicating that HSR found the local optimum routing scheme when considering the tradeoff.}
\end{figure}

%failure cases (Qwen-2B) in appendix

\begin{wrapfigure}{r}{0.5\textwidth}
\centering
\begin{tabular}{|p{1.5cm}||p{1.2cm}|p{1.2cm}|p{1.2cm}|}
\hline
Base Model & Prop & Pred & Counter\\
\hline
Gemini-3-Pro & 28\% & 23\% & 48\% \\
\hline
Qwen3-VL-4B-Thinking & 27\% & 36\% & 38\%\\
\hline
\end{tabular}
\label{table:Percentage}
\caption{The percentage of questions assigned to simulation. Two models were evaluated as the base model for HSR: Gemini-3-Pro and Qwen3-VL-4B-Thinking. Three question categories were evaluated, including property (Prop), predictive (Pred), and counterfactual (Counter), across all physical domains. Both base model tends to assign less questions to simulation for questions in property category, but more questions to simulation for questions in counterfactual category.}
\end{wrapfigure}

\paragraph{HSR determines routing scheme based on accuracy and computational costs} Since our HSR model's core ability is to provide a routing scheme given a set of physical reasoning questions, it is necessary to analyze the criterion in which this scheme is determined. Looking back to section \ref{rational} and \ref{sec:model}, we analyze the relationship between accuracy and the computational cost. As stated in \ref{sec:model}, the computational cost is controlled by parameter $\tau_t$. But this parameter $\tau_t$ also translates to the number of questions being assigned to "simulation" method, given that simulation is more costly than Chain-of-Thought. Therefore, instead of analyzing the relationship between parameter $\tau_t$, we transform the parameter $\tau_t$ to "number of questions assigned to simulation". 

As figure \ref{fig:model-results} shows, we observe that the routing scheme determined by the HSR model is mathematically the local maximum of the function that describe the relationship between two variables: accuracy and number of questions assigned to simulation. This indicates that our HSR model does not always go for the highest accuracy, if the computational cost is extremely high, like assigning all questions to simulation. Although the HSR model prioritize accuracy, instead, it optimizes the routing scheme based on the tradeoff between accuracy and computational costs. which is indicated by the qualitative results that our HSR model is able to reach local maximum.

\paragraph{HSR picks up categorical difference without training label}

\label{HSRcategorical}

As we shown in section \ref{category-difference}, simulation only and chain-of-thought only achieved different accuracy when the question category changes, where Chain-of-thought only performed better than simulation only in property category; simulation only perform better than chain-of-thought only in counterfactual category; simulation only and chain-of-thought performed similarly well in predictive category. We wonder if our HSR model picked up such categorical difference in each method. 

As shown in table \ref{table:Percentage}, we analyzed the percentage of questions assigned to simulation for each question category, collasping across the two physical domains. Regardness of base VLMs, the HSR tends to route property questions to Chain-of-thought more, indicated by the fact that only 28\% of the questions were routed to simulation for property category. But the reverse is true for the counterfactual category, where the HSR model tends to route large proportion ($\sim 40-50\%$) of questions to simulation. In the predictive category, when Gemini-3-Pro is set as the base model, less questions are assigned to simulation, comparing with the proportion of questions that are assigned to simulation in the predictive categories, when Qwen3-VL-4B-Thinking is set as the base model. 

Notice that the category label of the question were never passed into the model, and such difference in proportion of question assigned to simulation indicates that model picked up this feature without specific training label provided.

%% clarify the figure caption
%% Wrap figure

\begin{figure}[!tbp]
  \begin{minipage}[b]{0.33\textwidth}
    \includegraphics[trim=5 8 10 10,clip,width=\textwidth]{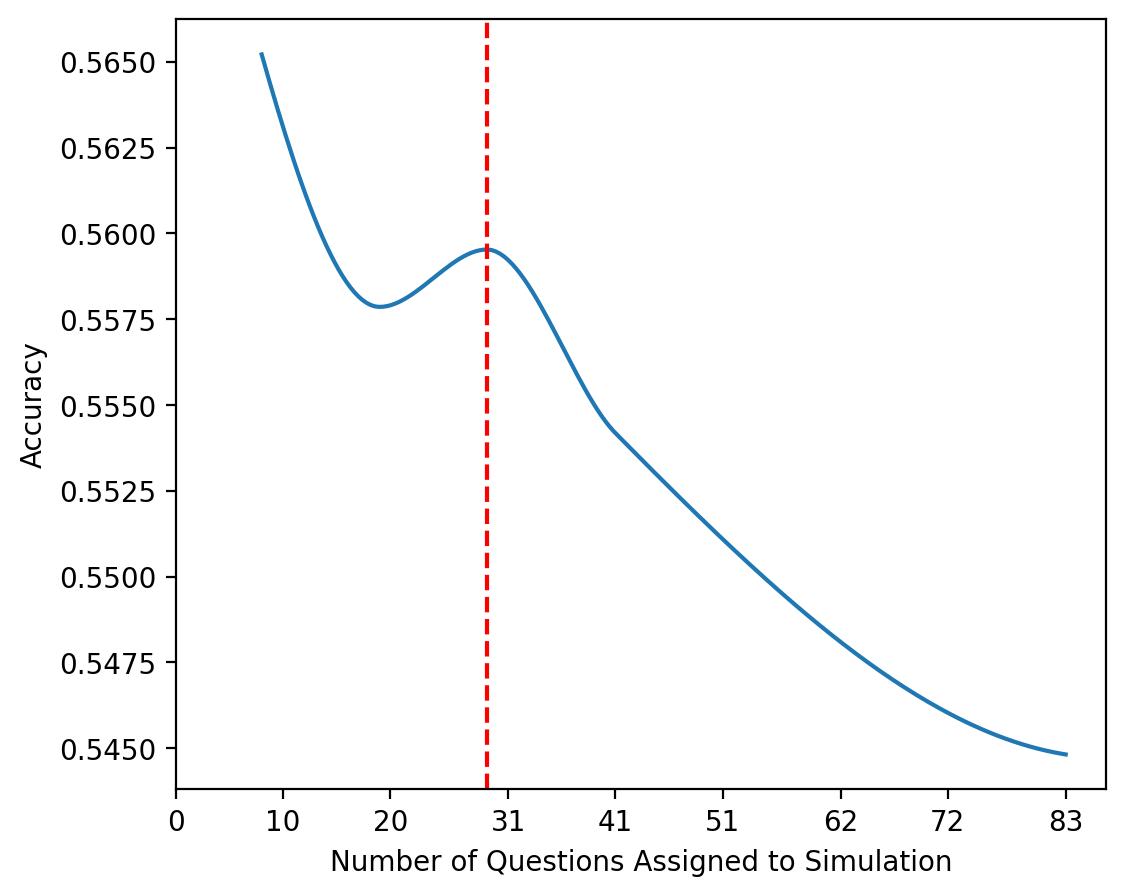}
  \end{minipage}
  \begin{minipage}[b]{0.33\textwidth}
    \includegraphics[trim=5 8 10 10,clip,width=\textwidth]{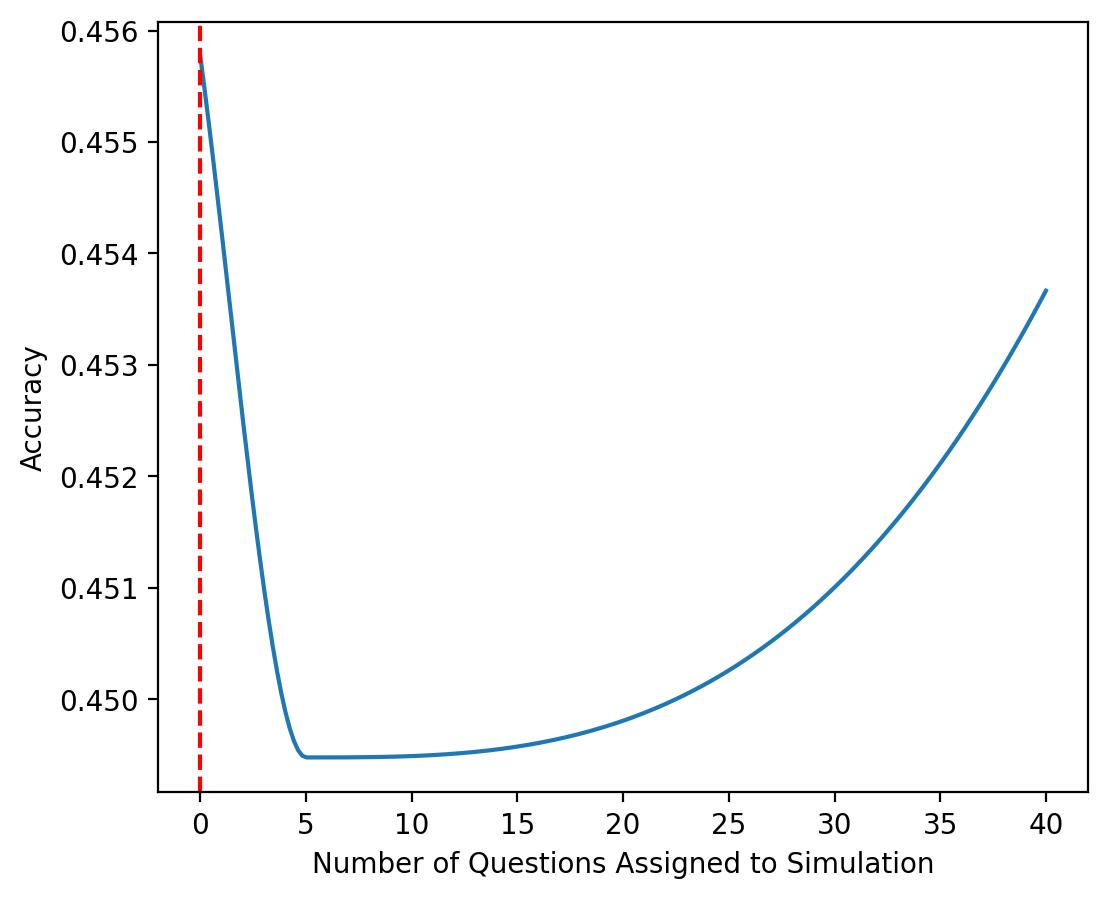}
  \end{minipage}
  \begin{minipage}[b]{0.33\textwidth}
    \includegraphics[trim=5 8 10 10,clip,width=\textwidth]{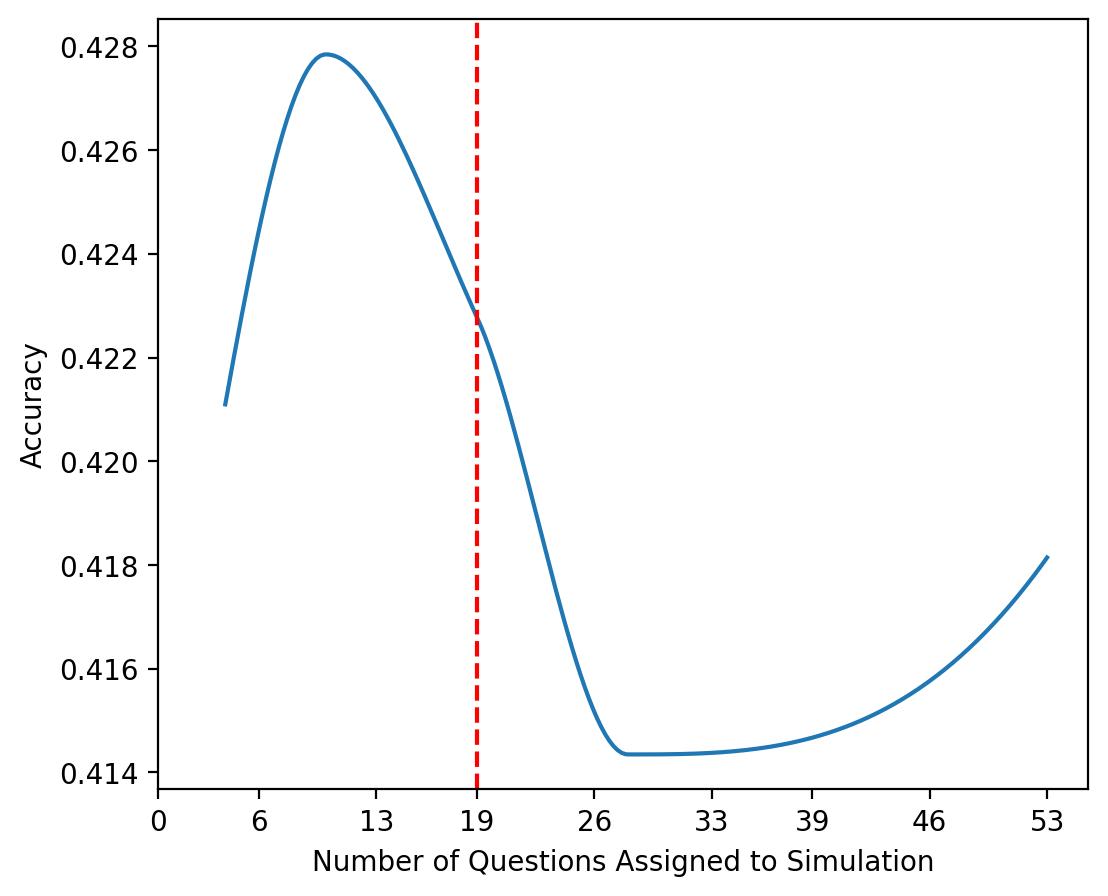}
  \end{minipage}
  \\[0.5em]
  % Second row
  \begin{minipage}[b]{0.33\textwidth}
    \includegraphics[trim=5 8 10 10,clip,width=\textwidth]{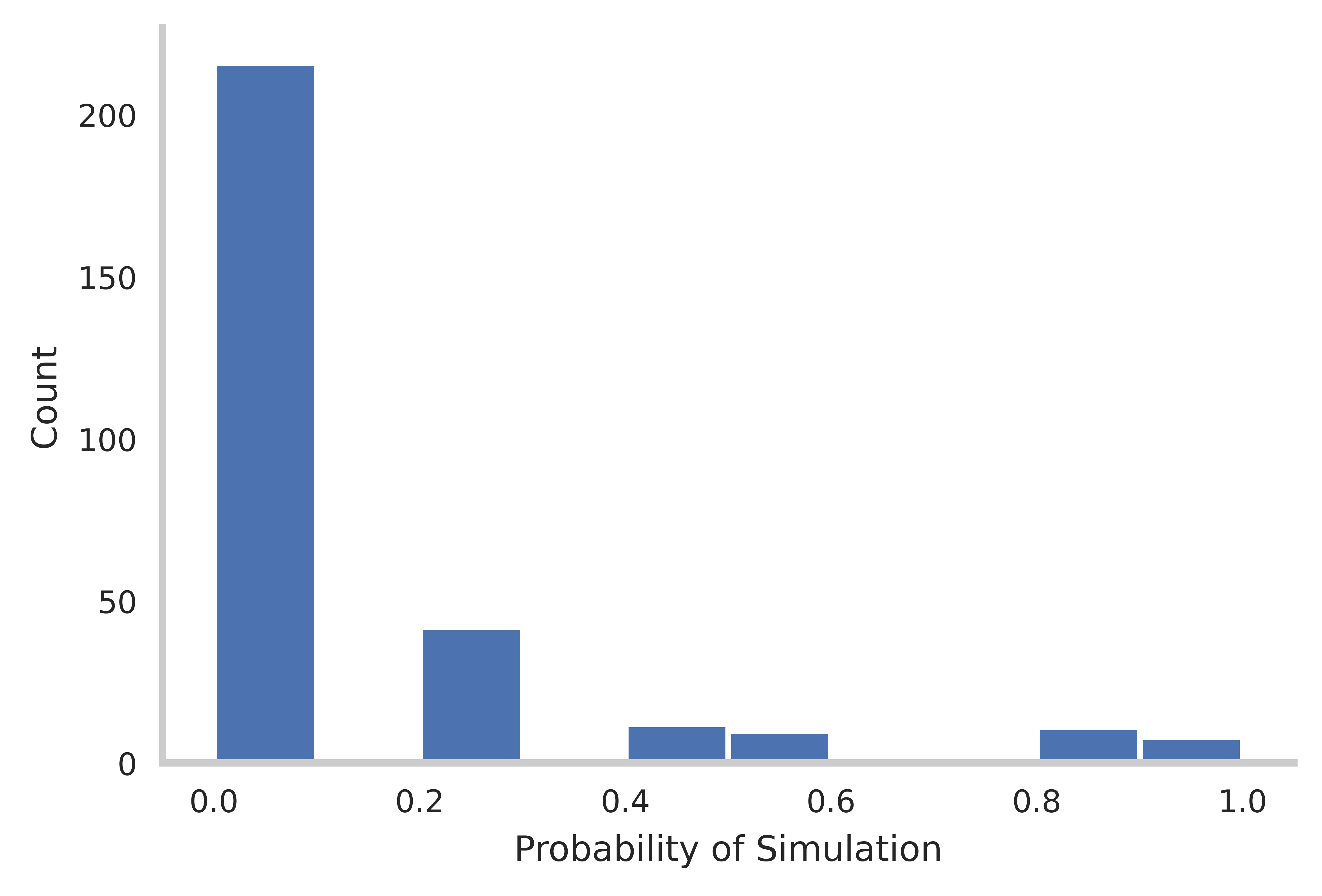}
  \end{minipage}
  \begin{minipage}[b]{0.33\textwidth}
    \includegraphics[trim=5 8 10 10,clip,width=\textwidth]{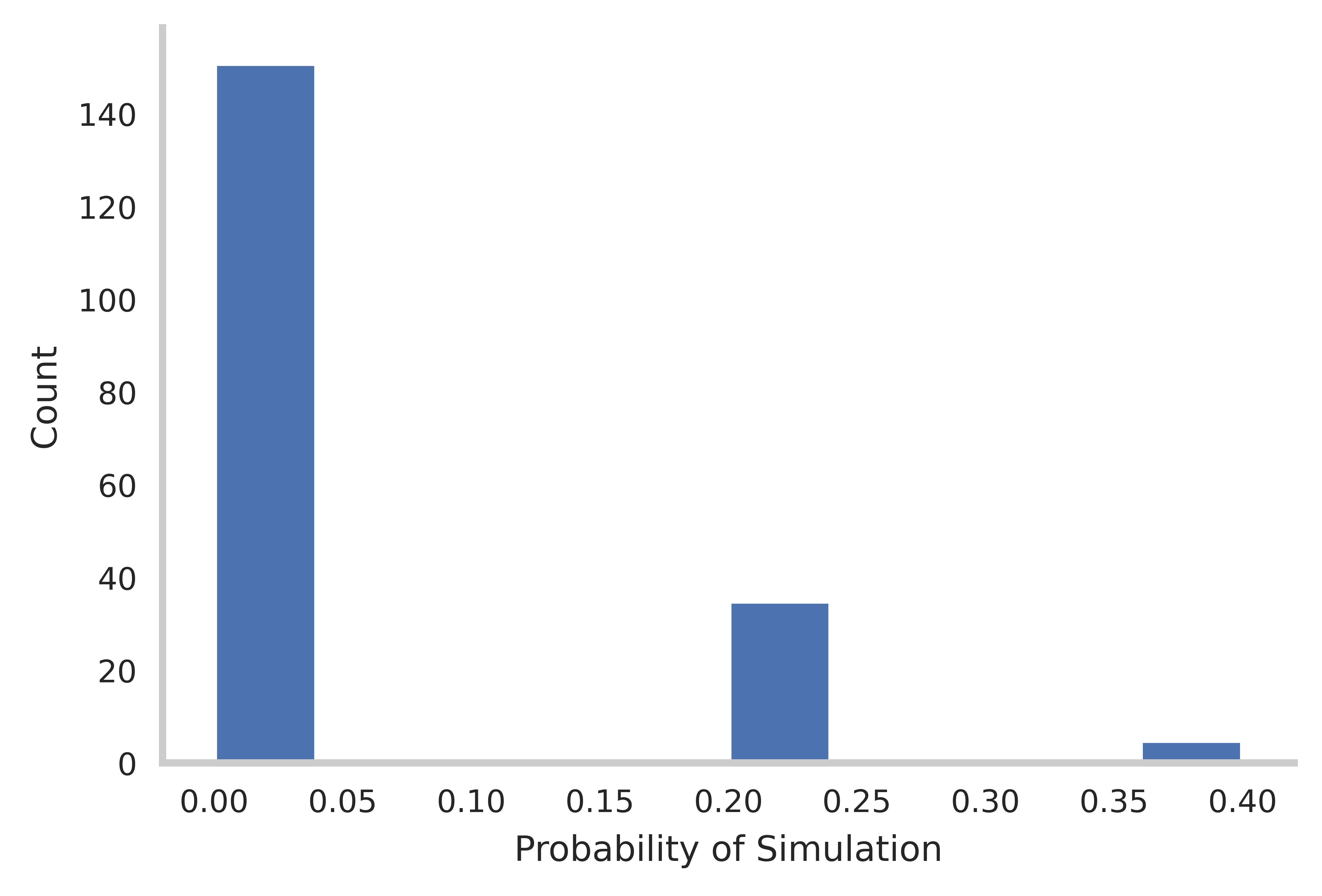}
  \end{minipage}
  \begin{minipage}[b]{0.33\textwidth}
    \includegraphics[trim=5 8 10 10,clip,width=\textwidth]{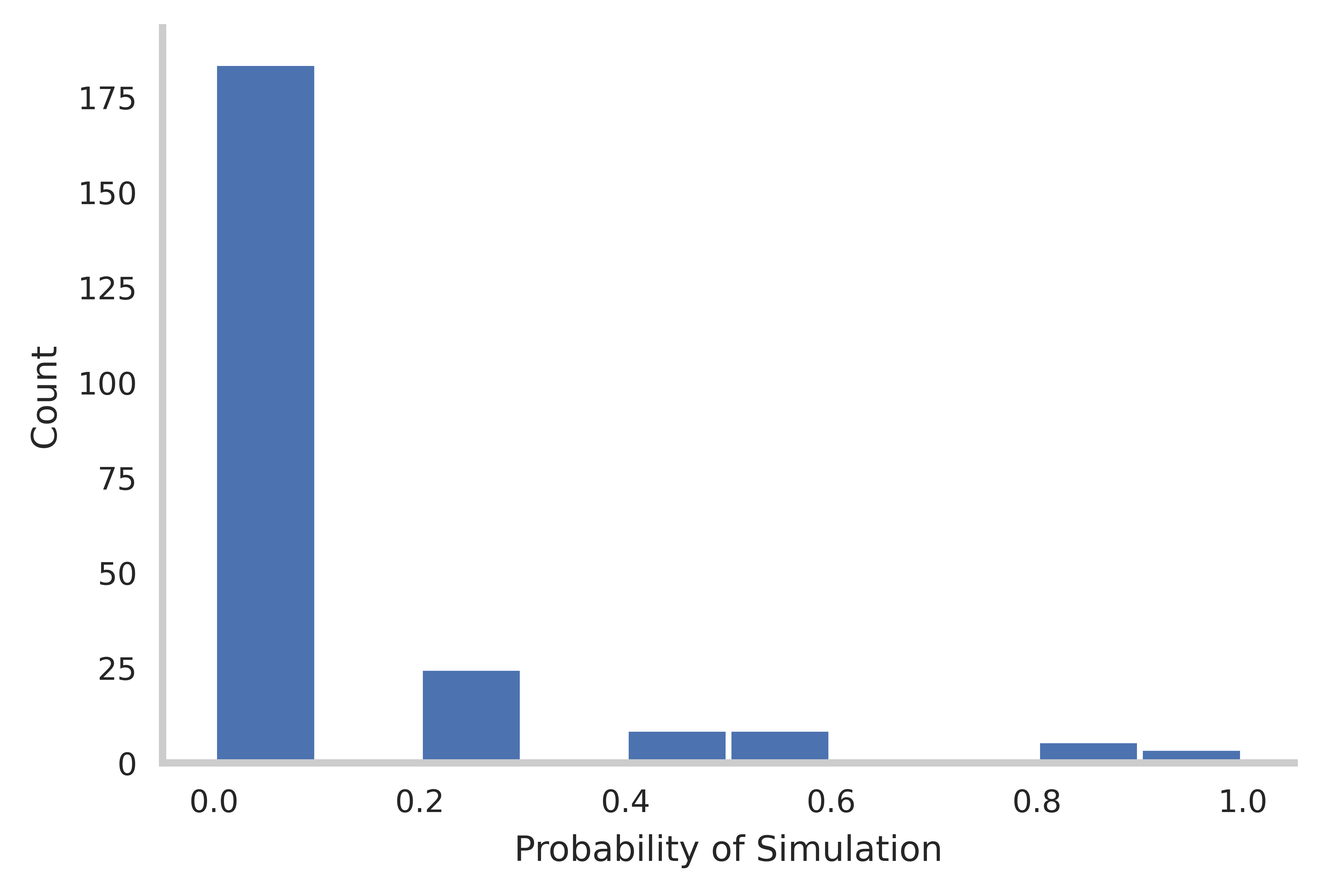}
  \end{minipage}
  \label{fig:zero-shot}
  \caption{\textbf{First row}: Accuracy vs Number of Questions Assigned to Simulation, when VLM-as-a-Judge serves as the simulation controller in the HSR model, with Gemini-3-Pro as the base model. The first column is the property category collasped across two physical domains. The second column is the predictive category collapsed across two physcial domains. The third column is the counterfactual category collapsed across two physical domains. The red dash line is the number of questions assigned to simulation, which is determined by VLM-as-a-Judge at threshold $\tau_t = 0.53$. The plot is showing uninterpretable results, showing the ineffectiveness of VLM-as-a-Judge as a simulator controller. \textbf{Second row:} Histogram of the probability of simulation predicted by VLM-as-a-Judge. The probability density concentrates around 0.}
\end{figure}

\subsection{Is the Simulation Controller Just Learning Trivial Text Feature About The Questions?}

As shown in section \ref{HSRcategorical}, our model is able to pick up the categorical difference without specific training labels, but one speculation still remains: is it possible that the simulator controller is picking up trivial text feature about the questions input into the simulation controller? Here we present an example to demonstrate that questions were assigned to simulation or chain-of-thought is not based on simple text features that could be picked up.

\begin{tcolorbox}[title=Counterfactual Question Example 1]
\textbf{Questions:}
"If the pink stick were removed, which container would light blue fluid flow into?
A. Green container
B. Blue container
C. Pink container
D. Brown container"\\
\textbf{Decision:} Simulation
\end{tcolorbox}

\begin{tcolorbox}[title=Counterfactual Question Example 2]
\textbf{Questions:}
"If the orange stick were removed, which container would light blue fluid flow into?
A. Green container
B. Pink container
C. Brown container
D. Purple container"\\
\textbf{Decision:} Chain-of-thought
\end{tcolorbox}

From the examples shown above, we can see that both questions are structurally similar, given that counterfactual condition like "if the pink stick were removed" is an obvious feature in the question. But two questions were routed into two different methods. Therefore, the simulation controller is not just learning trivial simple text features when deciding the routing scheme.

\subsection{Can VLM-as-a-Judge Act as a Zero-Shot Simulation Controller?}

The simulator controller in our HSR model serves as the most important component. but a possibility remains that we could potentially replacing this simulator controller with a VLM-as-a-Judge method, eliminating the necessity of training a separate MLP to serve as the controller. 

To test this possibility, we implemented VLM-as-a-Judge \citep{lee2024prometheus} method to replace the simulator controller (MLP) in our HSR model. The prompt we used for VLM-as-a-Judge is in Appendix \ref{prompt}. We used Gemini-3-Pro as the VLM in the VLM-as-a-Judge and as the base VLM in the HSR model, since Gemini-3-Pro model is the most powerful model among the models we used. As shown in the figure \ref{fig:zero-shot}, by replacing the MLP simulator controller with a VLM-as-a-Judge, the HSR model failed to produce effective routing scheme, indicated by the uninterpretable plot showing the relationship between accuracy and number of questions assigned to simulation.

We investigated into the reason behind such uninterpretable plot, and found that VLM-as-a-Judge cannot reliably judge if a question should be routed to Chain-of-thought or simulation, as indicated by the concentration of the probability of simulation around 0. This suggests that the VLM-as-a-Judge just monotonically assigned all questions to Chain-of-thought, rarely assigning questions to simulation, which clearly contradicts with our finding in section \ref{category-difference}.

\subsection{Smaller Model Could Serve as a Controller For Larger Models}

In our HSR model, the simulation controller is trained based on training data generated by the base model. This raises an interesting observation that if a simulation controller trained based on training data generated by a small model could be applied to HSR model with base VLM that is a large model.

\begin{figure}[!tbp]
  \begin{minipage}[b]{0.33\textwidth}
    \includegraphics[trim=5 8 10 10,clip,width=\textwidth]{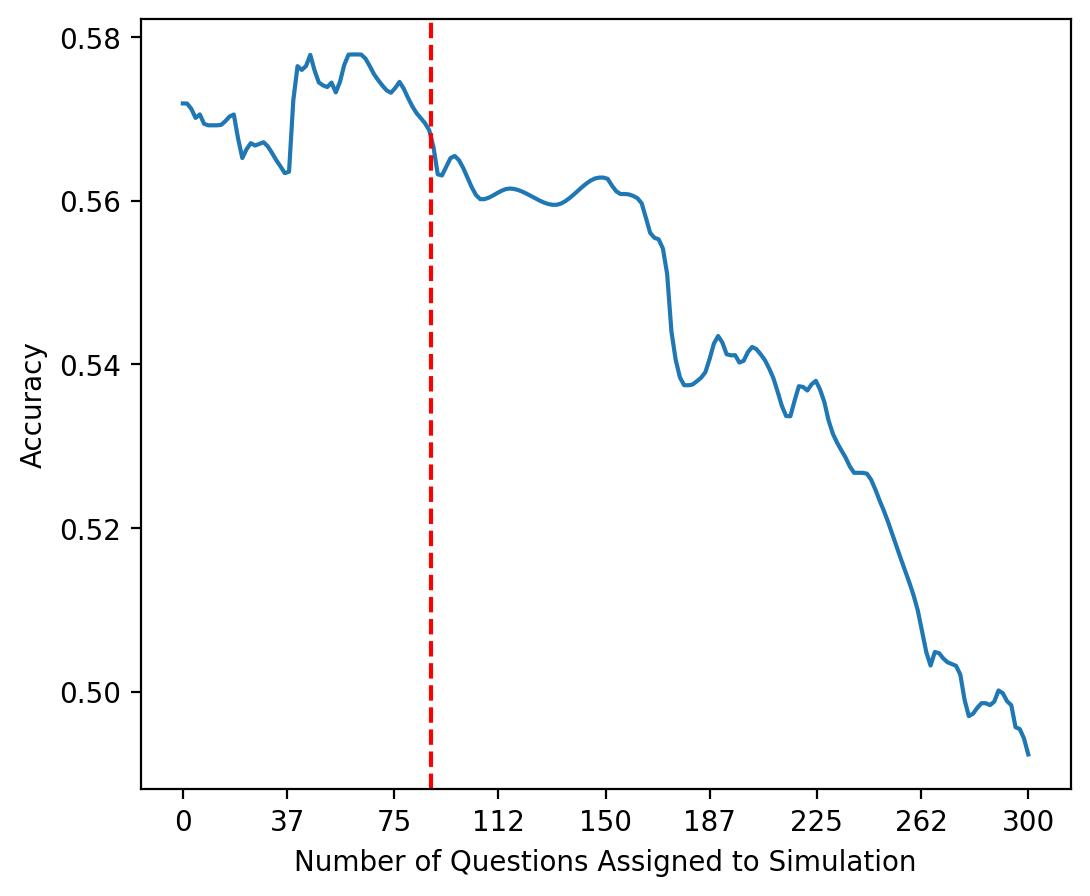}
  \end{minipage}
  \begin{minipage}[b]{0.33\textwidth}
    \includegraphics[trim=5 8 10 10,clip,width=\textwidth]{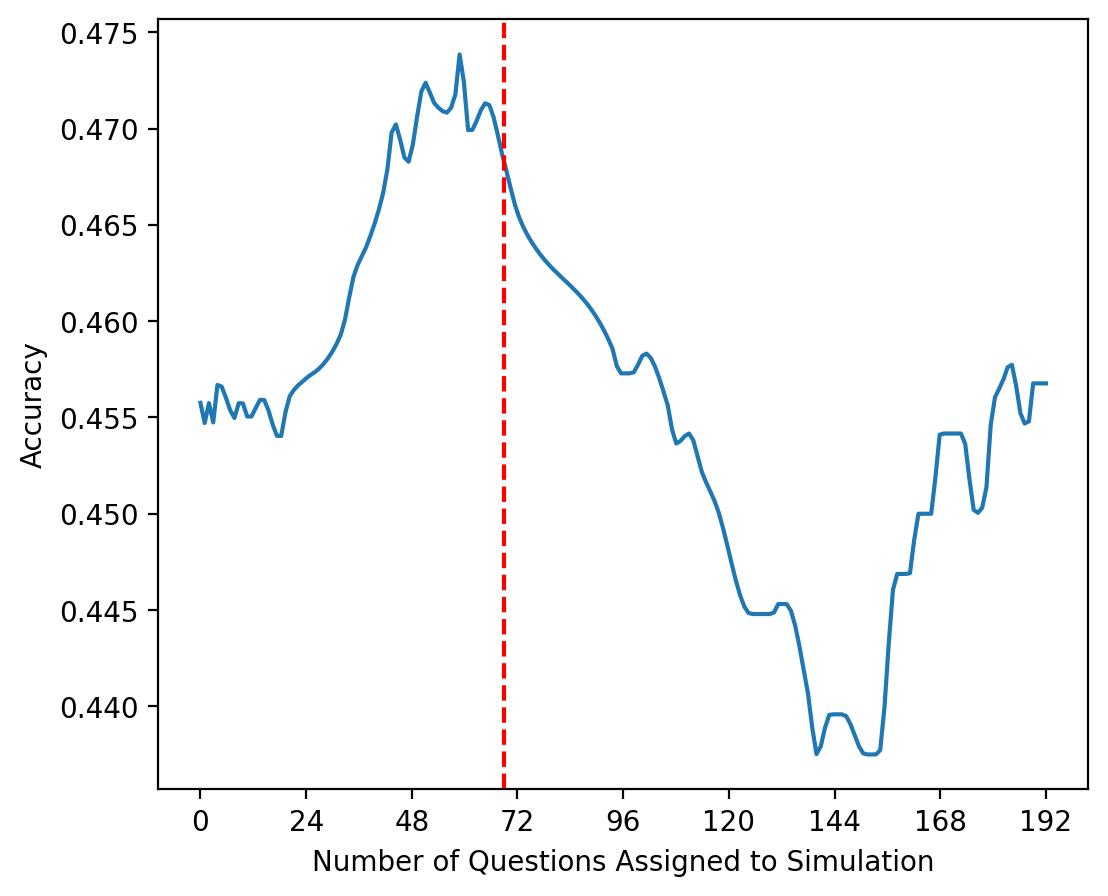}
  \end{minipage}
  \begin{minipage}[b]{0.33\textwidth}
    \includegraphics[trim=5 8 10 10,clip,width=\textwidth]{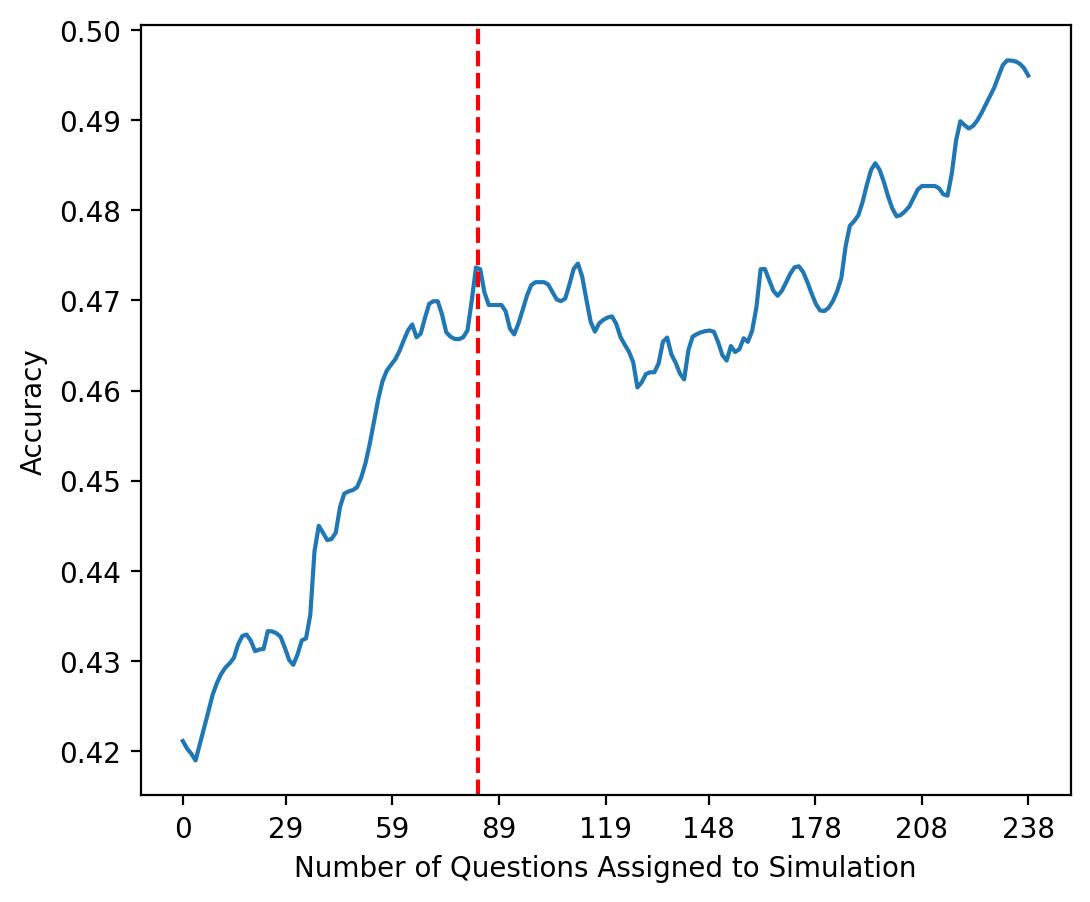}
  \end{minipage}
  \label{fig:smallmodel}
  \caption{Accuracy vs Number of Questions Assigned to Simulation. The HSR model with Gemini-3-Pro as the base VLM model and simulation controller trained based on Qwen3-VL-4B-Thinking is evaluated. The first column is the property category collasped across two physical domains. The second column is the predictive category collapsed across two physcial domains. The third column is the counterfactual category collapsed across two physical domains. The red dash line is the number of questions assigned to simulation, which is determined by HSR model at threshold $\tau_t = 0.53$. This dash line also indicates the local maximum of the function, indicating that HSR found the local optimum routing scheme when considering the tradeoff.}
\end{figure}

We conducted such experiment that we trained a simulation controller with training data generated by Qwen3-VL-4B-Thinking, but set the base VLM as Gemini-3-Pro. As shown in the figure \ref{fig:smallmodel}, we see that the simulation controller is able to again find the best routing scheme based on the tradeoff between accuracy and computational cost, despite being trained on training data generated by Qwen3-VL-4B-Thinking that is a way smaller model compared with Gemini-3-Pro. Such finding serves as an evidence that smaller model could be used to train a simulation controller that orchestrates a model that is way larger than itself, alike the idea demonstrated in papers like \citep{nielsen2025learning}.

%% fugu paper citation

\section{Conclusion}

In this paper, we introduce Hybrid Simulation and Reasoning Model (HSR) for physical reasoning in MLLMs. Our model is able to decide the appropriate method to solve a physical reasoning problem at test time, and finds an optimal balance between accuracy and cost. Further, our model can be trained on a smaller vision-language model that is only 4B parameters and deployed to improve performance on much bigger base models with $>$3T parameters like Gemini 3 Pro. 

We recognize that our HSR model is limited by the base model performance. But with the development of video models and VLMs, indicated by works like \citep{wiedemer2025video} and \citep{cosmos_cookbook_av_captioning_vqa_2026}, we are optimistic that our HSR model will become more powerful as the base model's ability improves in the future. And as the base model's ability improves, the routing scheme provided by HSR could be beneficial. For future work, we want to study the relationship between scene complexity and the necessity for simulation.

\bibliography{ref}

@article{smith2023integrating,
  title={Integrating heuristic and simulation-based reasoning in intuitive physics},
  author={Smith, Kevin and Battaglia, Peter and Tenenbaum, Joshua},
  year={2023},
  journal={PsyArXiv}
}

@article{liu2022mind,
  title={Mind's eye: Grounded language model reasoning through simulation},
  author={Liu, Ruibo and Wei, Jason and Gu, Shixiang Shane and Wu, Te-Yen and Vosoughi, Soroush and Cui, Claire and Zhou, Denny and Dai, Andrew M},
  journal={arXiv preprint arXiv:2210.05359},
  year={2022}
}

@article{wiedemer2025video,
  title={Video models are zero-shot learners and reasoners},
  author={Wiedemer, Thadd{\"a}us and Li, Yuxuan and Vicol, Paul and Gu, Shixiang Shane and Matarese, Nick and Swersky, Kevin and Kim, Been and Jaini, Priyank and Geirhos, Robert},
  journal={arXiv preprint arXiv:2509.20328},
  year={2025}
}

@article{kojima2022large,
  title={Large language models are zero-shot reasoners},
  author={Kojima, Takeshi and Gu, Shixiang Shane and Reid, Machel and Matsuo, Yutaka and Iwasawa, Yusuke},
  journal={Advances in neural information processing systems},
  volume={35},
  pages={22199--22213},
  year={2022}
}

@article{wei2022chain,
  title={Chain-of-thought prompting elicits reasoning in large language models},
  author={Wei, Jason and Wang, Xuezhi and Schuurmans, Dale and Bosma, Maarten and Xia, Fei and Chi, Ed and Le, Quoc V and Zhou, Denny and others},
  journal={Advances in neural information processing systems},
  volume={35},
  pages={24824--24837},
  year={2022}
}

@article{sosa2025blending,
  title={Blending simulation and abstraction for physical reasoning},
  author={Sosa, Felix A and Gershman, Samuel J and Ullman, Tomer D},
  journal={Cognition},
  volume={254},
  pages={105995},
  year={2025},
  publisher={Elsevier}
}

@inproceedings{zheng2024contphy,
  title={ContPhy: Continuum Physical Concept Learning and Reasoning from Videos},
  author={Zheng, Zhicheng and Yan, Xin and Chen, Zhenfang and Wang, Jingzhou and Lim, Qin Zhi Eddie and Tenenbaum, Joshua B and Gan, Chuang},
  booktitle={International Conference on Machine Learning},
  year={2024},
  organization={PMLR}
}

@article{liu2024mind,
  title={Mind your step (by step): Chain-of-thought can reduce performance on tasks where thinking makes humans worse},
  author={Liu, Ryan and Geng, Jiayi and Wu, Addison J and Sucholutsky, Ilia and Lombrozo, Tania and Griffiths, Thomas L},
  journal={arXiv preprint arXiv:2410.21333},
  year={2024}
}

@misc{cosmos_cookbook_av_captioning_vqa_2026,
  title={Post-train Cosmos Reason 2 for Autonomous Vehicle Video Captioning and VQA},
  author={Jin, Jingyi and Zhang, Shun and Yang, Xiaodong and Wang, Joseph},
  organization={NVIDIA and Uber},
  year={2026},
  month={January},
  howpublished={\url{https://nvidia-cosmos.github.io/cosmos-cookbook/recipes/post_training/reason2/video_caption_vqa/post_training.html}},
  note={NVIDIA Cosmos Cookbook}
}

@article{chow2025physbench,
  title={Physbench: Benchmarking and enhancing vision-language models for physical world understanding},
  author={Chow, Wei and Mao, Jiageng and Li, Boyi and Seita, Daniel and Guizilini, Vitor and Wang, Yue},
  journal={arXiv preprint arXiv:2501.16411},
  year={2025}
}

@article{ullman2017mind,
  title={Mind games: Game engines as an architecture for intuitive physics},
  author={Ullman, Tomer D and Spelke, Elizabeth and Battaglia, Peter and Tenenbaum, Joshua B},
  journal={Trends in cognitive sciences},
  volume={21},
  number={9},
  pages={649--665},
  year={2017},
  publisher={Elsevier}
}

@article{battaglia2013simulation,
  title={Simulation as an engine of physical scene understanding},
  author={Battaglia, Peter W and Hamrick, Jessica B and Tenenbaum, Joshua B},
  journal={Proceedings of the national academy of sciences},
  volume={110},
  number={45},
  pages={18327--18332},
  year={2013},
  publisher={National Academy of Sciences}
}

@article{hartshorne2025insights,
  title={Insights into cognitive mechanics from education, developmental psychology and cognitive science},
  author={Hartshorne, Joshua K and Jing, Mengguo},
  journal={Nature Reviews Psychology},
  pages={1--15},
  year={2025},
  publisher={Nature Publishing Group US New York}
}

@article{Radfordetal2021,
  author  = {Radford, Alec and Kim, Jong Wook and Hallacy, Chris and Ramesh, Aditya and Goh, Gabriel and Agarwal, Sandhini and Sastry, Girish and Askell, Amanda and Mishkin, Pamela and Clark, Jack and Krueger, Gretchen and Sutskever, Ilya},
  title   = {Learning Transferable Visual Models From Natural Language Supervision},
  journal = {Proceedings of the 38th International Conference on Machine Learning},
  year    = {2021},
  volume  = {139},
  pages   = {8748--8763},
  doi     = {10.48550/arxiv.2103.00020}
}

@inproceedings{reimers-2019-sentence-bert,
    title = "Sentence-BERT: Sentence Embeddings using Siamese BERT-Networks",
    author = "Reimers, Nils and Gurevych, Iryna",
    booktitle = "Proceedings of the 2019 Conference on Empirical Methods in Natural Language Processing",
    month = "11",
    year = "2019",
    publisher = "Association for Computational Linguistics",
    url = "https://arxiv.org/abs/1908.10084",
}

@article{hacohen2024ltx,
  title={Ltx-video: Realtime video latent diffusion},
  author={HaCohen, Yoav and Chiprut, Nisan and Brazowski, Benny and Shalem, Daniel and Moshe, Dudu and Richardson, Eitan and Levin, Eran and Shiran, Guy and Zabari, Nir and Gordon, Ori and others},
  journal={arXiv preprint arXiv:2501.00103},
  year={2024}
}

@article{team2023gemini,
  title={Gemini: a family of highly capable multimodal models},
  author={Team, Gemini and Anil, Rohan and Borgeaud, Sebastian and Alayrac, Jean-Baptiste and Yu, Jiahui and Soricut, Radu and Schalkwyk, Johan and Dai, Andrew M and Hauth, Anja and Millican, Katie and others},
  journal={arXiv preprint arXiv:2312.11805},
  year={2023}
}

@misc{veorobotics2025,
      title={Evaluating Gemini Robotics Policies in a Veo World Simulator}, 
      author={Gemini Robotics Team and Krzysztof Choromanski and Coline Devin and Yilun Du and Debidatta Dwibedi and Ruiqi Gao and Abhishek Jindal and Thomas Kipf and Sean Kirmani and Isabel Leal and Fangchen Liu and Anirudha Majumdar and Andrew Marmon and Carolina Parada and Yulia Rubanova and Dhruv Shah and Vikas Sindhwani and Jie Tan and Fei Xia and Ted Xiao and Sherry Yang and Wenhao Yu and Allan Zhou},
      year={2025},
      eprint={2512.10675},
      archivePrefix={arXiv},
      primaryClass={cs.RO},
      url={https://arxiv.org/abs/2512.10675}, 
}

@inproceedings{lee2024prometheus,
  title={Prometheus-vision: Vision-language model as a judge for fine-grained evaluation},
  author={Lee, Seongyun and Kim, Seungone and Park, Sue and Kim, Geewook and Seo, Minjoon},
  booktitle={Findings of the Association for Computational Linguistics: ACL 2024},
  pages={11286--11315},
  year={2024}
}

@article{nielsen2025learning,
  title={Learning to Orchestrate Agents in Natural Language with the Conductor},
  author={Nielsen, Stefan and Cetin, Edoardo and Schwendeman, Peter and Sun, Qi and Xu, Jinglue and Tang, Yujin},
  journal={arXiv preprint arXiv:2512.04388},
  year={2025}
}

@article{yang2026ares,
  title={Ares: Adaptive Reasoning Effort Selection for Efficient LLM Agents},
  author={Yang, Jingbo and Hou, Bairu and Wei, Wei and Bao, Yujia and Chang, Shiyu},
  journal={arXiv preprint arXiv:2603.07915},
  year={2026}
}

@article{waheed2025less,
  title={Less is More Tokens: Efficient Math Reasoning via Difficulty-Aware Chain-of-Thought Distillation},
  author={Waheed, Abdul and Mitra, Chancharik and Wang, Laurie Z and Ramanan, Deva and Raj, Bhiksha},
  journal={arXiv preprint arXiv:2509.05226},
  year={2025}
}

@article{bai2025qwen3,
  title={Qwen3-vl technical report},
  author={Bai, Shuai and Cai, Yuxuan and Chen, Ruizhe and Chen, Keqin and Chen, Xionghui and Cheng, Zesen and Deng, Lianghao and Ding, Wei and Gao, Chang and Ge, Chunjiang and others},
  journal={arXiv preprint arXiv:2511.21631},
  year={2025}
}
\bibliographystyle{plainnat}

\appendix
\section{Dataset Details}

\label{dataset}

We consider questions in three different categories: property, counterfactual, predictive and two physical domains: fluid hourglass and bouncing ball. These two sub-domains represent fluid dynamics and elastic collision, which are both non-rigid body motions.  

In total, there are 150 questions in property category for both sections; There are 104 questions in temporal predictive category and 116 questions in counterfactual category for fluid hourglass setting; There are 88 questions in temporal predictive category and 122 questions in counterfactual category for bouncing ball setting.

\section{Prompt}

\label{prompt}

\subsection{Prompt Used for Counterfactual Simulation Prior Generation}

Here we present the prompt that we used when implementing Gemini-3-Pro and Nano-Banana-Pro to prepare the physical prior of the counterfactual simulation for physical simulator (LTX-video).

Step 1: Judging the neediness for counterfactual simulation

\begin{tcolorbox}[colframe=blue, title=Prompt for judging the neediness for counterfactual simulation]
Imagine you are a physical simulation expert. Here are some videos or images together and a question based on these images or videos. the question is: {question}. You have to analyze the images or videos and the question. Do you think you need to reason counterfactually in order to answer the question based on the videos or images? Answer in 'Yes' or 'No' Only.
\end{tcolorbox}

Step 2: Judging the modifications for counterfactual simulation

\begin{tcolorbox}[colframe=blue, title=Prompt for judging the modifications for counterfactual simulation]
Imagine you are a physical simulation expert. Here are some videos or images together and a question based on these images or videos. The question is: {question}. Does the question suggested any counterfactual modifications we have to make to the objects and setup in the images and videos? If so, output the modifications you want to make only. If you are suggesting multiple modifications, separate them by a newline character. Example modifications can be: Remove object A, add object A, etc.
\end{tcolorbox}

Step 3: Modifying the original question into non-counterfactual version.

\begin{tcolorbox}[colframe=blue, title=Prompt for modifying the original question into non-counterfactual version]
Here you will be presented with a question with it's answer: {question}. Please remove the counterfactual condition in the original question provided to you, and output the question with the answer choices provided in the original question, if there are any, in the form that aligns with the input question I give to you only.
\end{tcolorbox}

\subsection{Prompt Used for VLM-as-a-Judge}

Here we present the prompt that we used when implementing VLM-as-a-Judge:

\begin{tcolorbox}[colframe=blue, title=Prompt for VLM-as-a-Judge]

First, you will see a set of images or videos, along with a physical reasoning question based on this set of images or videos. Then, the physical reasoning question is: {question}. Do you think in order to answer this question, you will need to perform extra simulation in a physics simulator, instead of using chain-of-thought only? Answer in 'Yes' or 'No' only."

\end{tcolorbox}

%prompt for vlm-as-a-judge
%results about Qwen-2B; Failure cases
%

\section{Additional Analysis}

%examples of routing
%

\label{additional}

\subsection{Impact of Base Model}

We conducted all of our experiment using various models including powerful closed-sourced model and open-source models. Given the importance of base VLM models in our HSR model, we tested the boundary that to which extent does HSR fails to work due to the incapability of the base VLM model. Therefore, we conducted experiment using Qwen3-VL-2B-Thinking, which is a even smaller model thant Qwen3-VL-4B-Thinking, and found that this model failed miserable as a base model in the HSR model. 

Therefore, here we present some of the failure cases in which Qwen3-VL-2B-Thinking struggles with basic question answering due to repetitiveness and congested Chain-of-thought.

\begin{tcolorbox}[colframe=blue, title=Example 1]

Question: "Is the density of blue fluid equal to that of pink fluid?"\\

Answer: "So, let's look at the video. The blue and pink fluids. \textbf{wait}, in the video, when the blue fluid is moving, does it interact with the pink? \textbf{Wait}, the question is about their densities. But in the simulation, maybe the colors are just visual. \textbf{Wait}, but the key here is to check if they have the same density. \textbf{Wait}, in the video, when the fluids are moving, maybe they're the same. \textbf{Wait}, but let's think. \textbf{Wait}, the problem is whether the density is equal. But in the video, maybe the blue and pink fluids are moving in a way that suggests same density. \textbf{Wait}, but maybe I need to check the motion. \textbf{Wait}, in the video, the blue fluid and pink fluid\u2014do they have the same density? \textbf{Wait}, maybe the answer is yes. \textbf{Wait}, but maybe not. \textbf{Wait}, let's see. \textbf{Wait}, the question is: Is the density of blue fluid equal to that of pink fluid? So, if in the simulation, the two fluids are moving in the same way, maybe they have same density. \textbf{Wait}, but maybe the answer is yes. \textbf{Wait}, but maybe I need to think again. \textbf{Wait}, the video shows the blue and pink fluids. \textbf{Wait}, in the simulation, maybe they have the same density. \textbf{Wait}, but the question is asking if they are equal. So, based on the video, if the blue and pink fluids are moving in the same direction or same way, maybe they have same density. \textbf{Wait}, but maybe the answer is yes. \textbf{Wait}, but I need to check. \textbf{Wait}, in the video, the blue fluid and pink fluid\u2014do they have the same density? \textbf{Wait}, maybe the answer is yes. \textbf{Wait}, but maybe not. \textbf{Wait}, perhaps the answer is no. \textbf{Wait}, but I need to think. \textbf{Wait}, the problem is that in the simulation, the blue and pink fluids might have different densities. .....\textbf{Wait} *stop"

\end{tcolorbox}

\begin{tcolorbox}[colframe=blue, title=Example 2]

Question: "Which container will the fluid from the other orange emitter flow into? A. Cyan container B. Pink container C. Gray container D. Green container"\\

Answer: ""Got it, let's look at the video. The orange emitters are the ones with orange colors at the top. The fluid from the orange emitters flows into the containers below. Let's check the containers: the bottom row has gray, green, cyan, pink. \textbf{Wait}, the video shows that the orange fluid flows into the blue container? \textbf{Wait} no, let's see. \textbf{Wait}, the containers are the ones at the bottom. Let's check the colors. The containers are: first one (leftmost) is gray, then green, then cyan, then pink. \textbf{Wait}, the orange emitter\u2014\textbf{Wait}, the emitters are the orange ones at the top. \textbf{Wait}, in the video, the orange fluid flows into the blue container? \textbf{Wait}, no. ..... \textbf{Wait}, no. \textbf{Wait}, maybe the answer is C. Gray container. \textbf{Wait}, no. \textbf{Wait}, I think the correct answer is C. Gray container. \textbf{Wait}, no. \textbf{Wait}, maybe the answer is C. Gray container. \textbf{Wait}, no. \textbf{Wait}, I'm confused. \textbf{Wait}, maybe the answer is C. Gray container. \textbf{Wait}, no. \textbf{Wait}, let's check the video again. The orange fluid flows into the blue container. \textbf{Wait}, no. \textbf{Wait}, the blue container is not there. \textbf{Wait}, the containers are the ones at the bottom. The first container is gray, then green, then cyan, then pink. \textbf{Wait}, the orange fluid flows into the blue container. \textbf{Wait}, no. \textbf{Wait}, maybe the orange fluid flows into the green container. \textbf{Wait}, no. \textbf{Wait}, the answer is C. Gray container? \textbf{Wait}, no. \textbf{Wait}, maybe the answer is C. Gray container. \textbf{Wait}, no. \textbf{Wait}, I think the correct answer is C. Gray container. \textbf{Wait}, no. \textbf{Wait}, maybe the answer is C. Gray container. \textbf{Wait}, no. \textbf{Wait}, I'm stuck. \textbf{Wait}, maybe the answer is C. Gray container. \textbf{Wait}, no. \textbf{Wait}, let's see: the video shows that the orange fluid flows into the blue container. \textbf{Wait}, no. \textbf{Wait}, the blue container is not there. \textbf{Wait}, the containers are the ones at the bottom. The first container is gray, then green, then cyan, then pink. \textbf{Wait}, the orange fluid flows into the blue container. \textbf{Wait}, no. \textbf{Wait}, maybe the orange fluid flows into the green container. \textbf{Wait}, no. \textbf{Wait}, the answer is C. Gray container? \textbf{Wait}, no. \textbf{Wait}, maybe the answer is C. Gray container. \textbf{Wait}, no. \textbf{Wait}, I think the correct answer is C. Gray container. \textbf{Wait},", *stop"

\end{tcolorbox}

\begin{tcolorbox}[colframe=blue, title=Example 3]

Question: "Is the plasticity of the orange ball much less than the blue ball?" \\

Answer: "So, let's analyze the video. The orange ball and blue ball are moving around. In the video, when the blue ball moves, it seems to interact with the orange ball. But \textbf{Wait}, the question is about plasticity. Plasticity here might refer to how much they deform or how they bounce. \textbf{Wait}, in the video, the orange ball and blue ball are rolling or moving. \textbf{Wait}, maybe the key is that the blue ball is moving more or less. \textbf{Wait}, no. \textbf{Wait}, the question is asking if the plasticity of the orange ball is much less than the blue ball. Plasticity is about how much a material deforms. If the blue ball is more deformable, then its plasticity is higher. \textbf{Wait}, but in the video, when the blue ball moves, does it deform? \textbf{Wait}, maybe the answer is based on whether they roll or bounce. \textbf{Wait}, the orange ball and blue ball are moving on the blocks. \textbf{Wait}, maybe the blue ball is more elastic or less plastic. \textbf{Wait}, but the question is about plasticity. Plasticity is the ability to deform permanently. If the blue ball is more plastic, then its plasticity is higher. \textbf{Wait}, but the question is asking if the orange ball's plasticity is much less than the blue ball's. So if the blue ball is more plastic, then yes. But how do we know from the video? \textbf{Wait}, maybe the blue ball is more likely to deform. .... \textbf{Wait}, maybe the answer is B. \textbf{Wait}, but I'm not sure. \textbf{Wait}, maybe the answer is B. \textbf{Wait}, but I think the answer is B. \textbf{Wait}, no. \textbf{Wait}, maybe the answer is B. \textbf{Wait}, but I'm not sure. \textbf{Wait}, maybe the answer is B. \textbf{Wait}, but I think the answer is B. \textbf{Wait}, no. \textbf{Wait}, maybe the answer is B. \textbf{Wait}, but I'm not sure. \textbf{Wait}, maybe the answer is B. \textbf{Wait}, but I think the answer is B. \textbf{Wait}, no. \textbf{Wait}, maybe the answer is B. \textbf{Wait}, but I" *stop

\label{appendix:qwen2b}

\end{tcolorbox}

\subsection{Reasoning Effort}

For Gemini-3-Pro, we notice that the model comes with the ability to adjust the reasoning effort during the reasoning process. Although when Gemini-3-Pro serves as the base VLM to support the Chain-of-thought method in our HSR model, the reasoning effort should by default be high, the role of reasoning effort when Gemini-3-Pro serves as the readout VLM in the simulation method remains unclear. Therefore, as an additional analysis, we tested the impact of reasoning effort when Gemini-3-Pro is set as the readout model of the simulation method in our HSR model.

According to our analysis in section \ref{result:HSR}, we focuses on two specific question categories: predictive and counterfactual, as these two categories are more relevant to the simulation method. As shown in the table \ref{table:low-effort}, we notice that the reasoning effort when Gemini-3-Pro is served as the readout VLM for the simulation method does does impact the final accuracy. Surprisingly, during the readout phase of simulation, higher reasoning effort did not increase the accuracy of the readout, instead it decreased the accuracy. 

\begin{table}[h!]
\centering
\begin{tabular}{|p{1.5cm}|p{1.2cm}||p{1cm}|p{1cm}||p{1cm}|p{1cm}|}
\hline
Base Model & Methods & FD-Pred & EC-Pred & FD-Counter & EC-Counter\\
\hline
\multirow{2}{1.5cm}{Gemini-3-Pro} & Sim-High & 0.315 & \textbf{0.570} & 0.435 & 0.440  \\
\cline{2-6}
& Sim-Low & \textbf{0.363} & 0.568 & \textbf{0.541} & \textbf{0.451} \\
\hline
\end{tabular}
\label{table:low-effort}
\caption{Evaluation results of simulation only (Sim) and chain-of-thought only (CoT) when Gemini-3-Pro set as the base model. We varied the reasoning effort in Gemini-3-Pro, which high reasoning effort (High) and low reasoning effort (Low) were used for evaluation. The evaluation is based on fluid hourglass (FD) and bouncing ball (EC) domains, with questions in predictive (Pred), and counterfactual (Counter) categories. The best accuracy in each question category and physical domains is highlighted in bold.}
\end{table}

\section{Statements}  ~\label{app:statements}

\subsection{Language Model Use}

LLMs were used in this work for literature review, editing the grammar error, and polishing writing in this paper. Also assisted the code writing.

\newpage
% Checklist omitted from the arXiv source build.

\end{document}